\pdfoutput=1
\documentclass[10pt,journal,compsoc]{IEEEtran}
\usepackage{booktabs}
\usepackage{color}
\usepackage{enumitem}
  
\ifCLASSOPTIONcompsoc
  \usepackage[nocompress]{cite}
\else
  \usepackage{cite}
\fi
\ifCLASSINFOpdf
  \usepackage[pdftex]{graphicx}
   \graphicspath{{../pdf/}{../jpeg/}}
   \DeclareGraphicsExtensions{.pdf,.jpeg,.png}
\else
   \usepackage[dvips]{graphicx}
   \graphicspath{{../eps/}}
  \DeclareGraphicsExtensions{.eps}
\fi
\usepackage{algorithm}
\usepackage{algorithmic}

\ifCLASSOPTIONcompsoc
  \usepackage[caption=false,font=footnotesize,labelfont=sf,textfont=sf]{subfig}
\else
  \usepackage[caption=false,font=footnotesize]{subfig}
\fi
\usepackage{fixltx2e}
\usepackage{url}
\begin{document}
\title{Supervised Anomaly Detection via Conditional Generative Adversarial Network and Ensemble Active Learning}

\author{Zhi Chen,
        Jiang Duan*,
        Li Kang,
        and~Guoping~Qiu% <-this % stops a space
\IEEEcompsocitemizethanks{\IEEEcompsocthanksitem  Zhi Chen, Jiang Duan and Li Kang are with the School of Economic Information Engineering, Southwestern University of Finance and Economics, Chengdu, China, 611130. Jiang Duan is the corresponding author\protect\\
	E-mail: $\{ chenzhi, duanj\_t, kangli\}$@swufe.edu.cn
	%E-mail: duanj\_t@swufe.edu.cn
	\IEEEcompsocthanksitem Guoping Qiu is with the College of Electronics and Information Engineering, Guangdong Key Laboratory for Intelligent Information Processing, and Shenzhen Institute of Artificial intelligence and Robotics for Society, Shenzhen University, Shenzhen 518060, China, and also with the School of Computer Science, University of Nottingham, Nottingham NG8 1BB, UK. \protect\\
    E-mail: guoping.qiu@nottingham.ac.uk
}

\thanks{}}


\markboth{}%
{Shell \MakeLowercase{\textit{Zhi Chen et al.}}: Supervised Anomaly Detection via Conditional Generative Adversarial Network and Ensemble Active Learning}



\IEEEtitleabstractindextext{%
\begin{abstract}
Anomaly detection has wide applications in machine intelligence but is still a difficult unsolved problem. Major challenges include the rarity of labeled anomalies and it is a class highly imbalanced problem. Traditional unsupervised anomaly detectors are suboptimal while supervised models can easily make biased predictions towards normal data. In this paper, we present a new supervised anomaly detector through introducing the novel Ensemble Active Learning Generative Adversarial Network (EAL-GAN). EAL-GAN is a conditional GAN having a unique one generator \emph{vs.} multiple discriminators architecture where anomaly detection is implemented by an auxiliary classifier of the discriminator. In addition to using the conditional GAN to generate class balanced supplementary training data, an innovative ensemble learning loss function ensuring each discriminator makes up for the deficiencies of the others is designed to overcome the class imbalanced problem, and an active learning algorithm is introduced to significantly reduce the cost of labeling real-world data. We present extensive experimental results to demonstrate that the new anomaly detector consistently outperforms a variety of SOTA methods by significant margins. The codes are available on Github.
\end{abstract}

}

\maketitle

\IEEEdisplaynontitleabstractindextext

% \fi
\IEEEpeerreviewmaketitle

\IEEEraisesectionheading{\section{Introduction}\label{sec:introduction}}

\IEEEPARstart{A}{nomalies} nomalies (also known as outliers, novelties, or faults) refer to the data points that deviate significantly from the majority of available data \cite{RN1}. Due to the valid, interesting and potentially valuable patterns they often represent, detecting such data could provide valuable knowledge in various real-world applications, such as finance fraud detection \cite{RN2}, intrusion detection \cite{RN3}, rare information detection in healthcare \cite{RN4},\cite{RN5} and video analysis \cite{RN6},\cite{RN7}. Three substantial challenges in these applications are: (i) It is very difficult to obtain sufficient anomalies and their ground-truths to train an anomaly detector due to the prohibitive cost of collecting and labeling such data. (ii) Anomalies often exhibit very different anomalous behaviors, as a result, training an anomaly detector could be quite a challenge for the commonly-used optimization techniques, which generally assume that data within the same class should be similar to each other \cite{RN8}. (iii) The available data could present a highly skewed distribution due to the rarity of anomalies. Consequently, most supervised models can easily make biased predictions to the normal data.

To address these challenges, researchers generally adopt a two-step unsupervised learning procedure to isolate the anomalies from normal data: They first learn to represent all the available data with new representations, e.g., distance metric spaces in \cite{RN9},\cite{RN10},\cite{RN11}, representations in a projected space \cite{RN12},\cite{RN13},\cite{RN14},\cite{RN15},\cite{RN16}, or latent spaces in generative adversarial networks (GANs) \cite{RN4},\cite{RN17},\cite{RN18},\cite{RN19}. And then, the data which significantly deviates from the established normal profiles will be identified as potential anomalies. In most of these unsupervised methods, representation learning and anomaly detector training are separated into two independent procedures, therefore, such methods may yield suboptimal representations or representations irrelevant to the anomaly detection task \cite{RN8}. To avoid this problem, some works have incorporated traditional anomaly scoring metrics into the representation learning objective to improve the quality of learned representations \cite{RN20}. However, these methods mainly focus on unsupervised solutions, so they share the common shortcoming of unsupervised learning, often identify anomalies that are merely uninteresting data or noise. In recent years, deep learning has gained remarkable success in various applications and has also shown promising potential in anomaly detection. However, most existing deep anomaly detectors \cite{RN21},\cite{RN22},\cite{RN23} are unsupervised methods, therefore they also suffer from the inherent limitations of such methods. 

Conditional generative adversarial networks (cGANs) \cite{RN24},\cite{RN25} have demonstrated exceptional ability in generating labeled samples indistinguishable from real world data. It therefore offers an appealing prospect of overcoming the problem of lack of labelled data in supervised anomaly detection by conditionally generating class balanced data. However, despite the promising performance of cGANs in modeling various conditional distributions, the performances of standard cGANs heavily depends on the size of labeled training data \cite{RN26}. Very importantly, a standard cGAN is never designed for the class imbalanced scenarios. In anomaly detection, labeled anomalies are rarities and the distribution of the available data is highly skewed. Therefore anomaly detection presents major challenges for a common cGAN.

In this paper, we present a supervised anomaly detection method through designing a cGAN featuring an ensemble of discriminators that are trained based on a novel ensemble active learning strategy. At the heart of our new method is the ensemble active learning generative adversarial network (EAL-GAN) which has the following novel features and advantages:

\begin{itemize}[topsep=0pt]
	\renewcommand{\labelitemi}{$\vcenter{\hbox{\small$\bullet$}}$}
	\setlength{\itemsep}{0pt}
	\setlength{\parsep}{0pt}
	\setlength{\parskip}{0pt}
	
	\item 
	\emph{Unique Network Architecture}. EAL-GAN is a conditional GAN featuring a unique one generator \emph{vs.} multiple discriminators architecture. Each discriminator consists of two classifiers simultaneously performs two classification tasks: (i) An adversarial classifier performs real and generated data classification, and (ii) an auxiliary classifier performs anomaly and normal data classification. An ensemble learning algorithm is developed to train the multiple discriminators to overcome the class imbalanced issue. The auxiliary classifier acts as an active learning sampler as well as the anomaly detector.
	
	\item
	\emph{Novel Ensemble Learning Loss function}. A novel ensemble learning loss function is designed to ensure that the ensemble of discriminators complementing each other. Each discriminator adaptively focuses on the samples wrongly classified by the others thus overcoming the problem of bias classification towards the classes with larger number of samples.
	
	\item
	\emph{Active Ensemble Learning Reduces the Cost of Labeling Data}. An active sampling strategy is incorporated into the EAL-GAN framework. The active sampling strategy progressively selects the data carrying the most information but accounting for relatively small proportion of the available real data to not only optimize the discriminator ensemble, but also guide the generator to produce sufficiently balanced fake data. The selected real data and the generated fake data are used to train the discriminators thus significantly reducing the cost of annotating anomaly data for training a fully supervised detector. Empirical results show that in a batch-wise training procedure, EAL-GAN can provide state-of-the-art performance by only labeling 5\% of the real data in each batch.
	
	\item
	\emph{Multi-Discriminator Ensemble Learning Enhances Regularization and Generalization}. In the proposed EAL-GAN framework, the discriminator is designed as a multi-task neural network. By effectively incorporating multiple discriminators into one ensemble with a novel ensemble learning loss function, the EAL-GAN shows enhanced capability to resist overfitting. This makes it easily to obtain an anomaly detector with optimal generalization or determine the stop node of GAN training by choosing the model with the best empirical performance (please see Fig. 2 and Section 3.3)
	
	\item 
	\emph{The EAL-GAN is a High Quality cGAN}. The unique architecture and innovative ensemble learning algorithm have ensured that EAL-GAN is a high-quality conditional data generator. This is demonstrated by the fact that anomaly detectors trained with only the generated data can be as good as those trained with the real data and that mixing generated data with the real data can further improve performances. 
	
	\item 
	\emph{State of the Art Performances}. Extensive experiments have been carried out on 20 widely used real world anomaly detection benchmark datasets and a variety of synthetic datasets. Performances are compared to 9 state-of-the-art anomaly detection methods from 6 categories in the literature. The new EAL-GAN method consistently outperforms the best methods available, often by significant margins, achieving AUC performance improvements range from 5.7\% over the best deep learning detector, to 16.4\% over the best traditional detector, and to 16.8\% over the best ensemble detector; and achieving Gmean performance improvements range from 37.9\% over the best deep learning detector, to 49.2\% over the best ensemble detector, and to 52.5\% over the best traditional detector.

\end{itemize}

\section{Related Work}
In this section, we summarize the progress in anomaly detection and active learning, both of which are related to the work in this paper.

\subsection{Anomaly Detection}
During the past decades, various anomaly detection methods have been introduced. We roughly group all the methods into: traditional methods and deep methods. More comprehensive review can be found in \cite{RN1},\cite{RN27},\cite{RN28}.

\subsubsection{Traditional Anomaly Detection Methods}
A common assumption in traditional anomaly detection is that the anomalies are rare and diverse, which makes it infeasible to collect and label sufficiently large data to build discriminative models. Thus, researchers attempt to create a model representing the normal data and identify the data that deviates significantly from the established normal behaviors as anomalies. Among various methods, density-based \cite{RN29},\cite{RN30},\cite{RN31},\cite{RN32},\cite{RN33}, clustering-based \cite{RN34},\cite{RN35}, distance-based methods \cite{RN9},\cite{RN10} and ensemble detectors \cite{RN36},\cite{RN37},\cite{RN38},\cite{RN39} are the most popular methods.

Density-based methods usually estimate the density function of the data, and then identify the anomalies as those having large deviations from the density peaks. For example, given a training set, the Gaussian mixture model (GMM) \cite{RN29} fits a given number of Gaussian distributions to the data via the Expectation-Maximization (EM)\cite{RN30} algorithm. The data points with the smallest likelihood are identified as anomalies. The Kernel Density Estimation (KDE) method \cite{RN31} approximates a dataset’s local density function by building a kernel function for each data point and then summing the local contributions of each kernel. Anomalies are identified as those which have significantly different local densities from their neighbors. KDE is efficient for the anomaly detection. However, as shown in \cite{RN32}, the original KDE is sensitive to outliers. To overcome this, recent studies have adopted robust loss function \cite{RN32} and weighted neighborhood density estimation \cite{RN33} to estimate the local density. The Support Vector Data Description (SVDD) \cite{RN12} does not explicitly define a density function, instead, it finds the smallest hypersphere in the projected high-dimensional space that can enclose all the normal data. Any data located outside such a hypersphere is identified as an anomaly. SVDD is known for its excellence in handling small sample size, high dimensionality and non-linearity. However, setting optimal hyper-parameter is the key, which largely depends on the available expertise. The reconstruction-based methods, such as Principal Component Analysis (PCA) \cite{RN13},\cite{RN15} and Matrix Factorization (MF) \cite{RN40}, allow for the projection of original instance into a latent space, where instances are more separable, and then reconstruct the original instance from the latent space. The data with high reconstruction error will be identified as anomalies. 

Another widely used anomaly detection method, clustering-based method, detects anomalies after clustering the data. The data located far away from any cluster center, the data located in very sparse or small clusters \cite{RN34} will be identified as anomalies. Since clustering methods don’t consider the ground-truths, they are suitable for the scenarios where certain level of robustness is required. For example, Nouretdinov et al.\cite{RN34} combined \emph{k}-means clustering, Local Outlier Factor (LOF) and SVDD to detect the anomalies in a dataset with imperfect labels. Shi and Zhang \cite{RN35} adopted clustering to iteratively detect anomalies in multi-dimensional data. The general clustering-based methods build on the assumption that the normal data are dense enough, however, this assumption cannot hold in some complex distributions. Worse still, clustering techniques need to pre-determine the cluster types and numbers, which can be crucial and very difficult in practical applications. All these inherent limitations can lead to reduced performance on data with high dimensionality or intricate relations.

Unlike other methods, distance-based methods make no assumption about the data distribution, and only use the distance measure to identify anomalies. Typical distance-based methods include: (1) \emph{k}-nearest neighbor (\emph{k}NN) \cite{RN9} method, which measures the rarity of the data according the distance to their \emph{k} nearest neighbors. (2) Fast Angle-based Outlier Detection (FastABOD) \cite{RN10}, which quickly estimates the angular variation of each data point using the radius and variance of angles measured at each data point. 

Research efforts have also been devoted to designing various ensemble learning for anomaly detection. Compared with single detector, ensemble learning combines several diverse and complementary models. It has been shown in \cite{RN36},\cite{RN37},\cite{RN38},\cite{RN39} that building ensemble detectors can substantially improve the efficacy of the above traditional anomaly detectors. Among various anomaly ensembles, iForest \cite{RN36} is the most popular one which builds an ensemble of tree detectors by performing recursive random splits on feature values, hence generating trees capable of isolating any data point into some subspaces.

\subsubsection{Deep Anomaly Detection Methods}
In recent years, deep learning methods have shown exceptional ability for discovering anomalies in complex data distributions. Current popular deep anomaly detectors include deep auto-encoders \cite{RN21},\cite{RN22},\cite{RN23} and GAN-based methods \cite{RN4},\cite{RN17},\cite{RN18},\cite{RN19}. Deep auto-encoder adopts a bottleneck-like deep architecture to map the original data into a low-dimensional representation space, and then reconstruct the original data from the learned representation. In such a compression and decompression procedure, the data with higher reconstruction-error will be identified as anomalies. The idea of deep auto-encoder is very similar to the traditional reconstruction-based methods, i.e., PCA and MF. The major difference between these two methods is the greater power of auto-encoder in modeling complex data distributions \cite{RN17}. 

GAN can capture the data distribution of real data through an adversarial learning process. Specifically, GAN leverages a generator network to generate realistic data, and a discriminator network to distinguish the generated data from the real data. Two networks compete with each other until the whole system converge to the Nash equilibrium. In such a condition, the generator can map a latent vector to realistic data, while the discriminator can reconstruct the latent vector from the real data. Any data with higher reconstruction-error will be treated as anomalies. Compared to traditional shallow models, GAN can capture more complex feature interaction, therefore, increased efforts have been devoted to this emerging technique. For example, Schlegl et al. \cite{RN4} designed a deep convolutional generative adversarial network (AnoGAN) to learn a manifold of normal anatomical variability. AnoGAN can map medical images from image space to the latent space, such that abnormal markers in medical images can be discovered. Donahue et al. \cite{RN19} proposed a Bidirectional Generative Adversarial Network (BiGAN) to jointly train the mapping from image space to latent space and vice versa. BiGAN has also been used in \cite{RN18}, in which various methods have been proposed to stabilize the GAN training and bring significantly better performance. Despite their promising performance, all these methods are unsupervised methods which use GAN as a powerful feature extractor or re-constructor. These methods may lead to a common problem of unsupervised anomaly detection that many of the anomalies they identify are uninteresting. Liu et al. \cite{RN17} proposed an unsupervised GAN with multiple generators for outlier detection. The proposed GAN actively selects potential anomalies from a uniform reference distribution and trains the discriminator with the sampled data. In such a manner, the discriminator can be trained as a binary classifier for outlier detection.

\subsection{Active Learning}
The basic idea of active learning is to iteratively select and label the most useful unlabeled samples to boost the model. As a hot research topic, various active learning methods are designed with different formulas, and the organization of them are almost based on two criteria, i.e., the uncertainty \cite{RN26},\cite{RN41},\cite{RN42},\cite{RN43} and diversity \cite{RN44},\cite{RN45}. While uncertainty-based methods directly query the most uncertain samples whose prediction labels have low confidence, the diversity-based methods select the most dissimilar samples, thus the information redundancy among the samples can be reduced. These methods have been successfully applied to a variety of problems, such as activity recognition \cite{RN42}, image classification \cite{RN46}, outlier detection \cite{RN17} and social bot detection \cite{RN47}.

Despite its effectiveness of handling the lack of labeled data problem, active learning still has some issues in many real-world applications. For example, uncertainty-based methods exhibit limitations in removing the redundancy between samples. In response to these issues, researchers have proposed various active learning methods. For example, while Wang et al. \cite{RN43} evaluated the uncertainty of a sample based on the information extracted from multiple kernels, Hasan et al. \cite{RN42} designed a sample selection strategy that considers both the informativeness and the contextual information of the individual activity, such that the proposed active learning can be successfully applied to activity recognition. Li et al. \cite{RN41} approach the active learning and feature selection as a joint learning problem. Specifically, in the matrix decomposition process, the proposed method selects both the samples and features that can best reconstruct the original data. 

Recently, researchers have also applied active learning to deep learning, where large number of labeled data are needed. For example, some works apply active learning methods in deep image classification tasks \cite{RN46}. There are also a few methods using deep generative models, such as GANs, to generate or acquire more effective labeled data \cite{RN48},\cite{RN49}. While these methods use GANs to improve active learning, we employ active learning to reduce the labeling cost for training a cGAN. Moreover, we evaluate the uncertainty of a sample with the confidence provided by an ensemble model. Therefore, the bias caused by adopting single metric or single model can be potentially reduced.

\section{Anomaly Detection Using cGAN and Ensemble Active Learning}
In the following parts, the superscript ${r}$ and ${g}$ denote the real distribution and generative distribution, respectively. We first formally give the problem statement and challenges in this study, and then describe the details of the proposed framework.

\subsection{Problem Statement}
We aim to find a classification boundary that can effectively separate anomalies from normal data. Specifically, given a dataset of \emph{n} training samples ${{\cal X} = \{ {x_1},{x_2},...,{x_n}\} }$ with ${{x_i} \in {\mathcal{R}^d} }$ , and label ${{\cal Y} = \{ {y_1},{y_2},...,{y_n}\}} $  where label ${{y_i} = 1}$ indicates ${x_i}$ is an anomaly and ${{y_i} = 0}$ for a normal data, our goal is to learn an anomaly detection function ${\phi (x) \in \{ 0,1\}} $ that assigns a label of 1 to an anomaly and 0 to normal data. To this end, one can minimize the loss ${{{\rm{{\cal L}}}_\phi }{\rm{}}}$ for ${\phi (x)}$ as follows:

\begin{small}
\begin{equation}
{{\rm{{\cal L}}}_\phi }{\rm{}} = \! - \frac{1}{n}\sum\limits_{i = 1}^n {({C_1}{y_i}\log (\phi ({x_i}))\! +\! {C_0}(1 - {y_i})\log (1 - \phi ({x_i})))}
\end{equation}
\end{small}

Where ${C_1}$ and ${C_0}$ denote the misclassification costs for the anomaly and normal data, respectively. Given a large training set with sufficiently balanced class distribution, minimizing the loss function in Eq.(1) forces $\phi (x)$ to converge to two conditions: (i) The data of the same class converge to the class distribution with the same density peaks, and the data of different classes correspond to different peak positions. (ii) $\phi ({x_i}) > \phi ({x_j})$ if ${x_i}$ is an anomaly and ${x_j}$ is a normal data point. As such, the anomaly detection function $\phi (x)$ can be learned in a supervised manner to find the classification boundary that can effectively separate the anomaly from normal data in the prediction space.

However, learning guided by the loss function in Eq.(1) can encounter serious problem due to the skewed distribution in the training data. There are far fewer occurrences of anomaly data and it can be prohibitively expensive to accurately label them, this has led to the situation where anomaly data is seriously under-represented. For learning algorithms based on gradient descent, most of the gradients will be produced by the easily available normal data. Consequently, the model can easily make biased predictions towards the normal data. To build an effective supervised model for this task, the fundamental problem here is how to better represent the anomaly data. In this paper, a conditional Generative Adversarial Network consisting of an ensemble of active learning discriminators (EAL-GAN) is designed to tackle the aforementioned challenges of designing a supervised anomaly detector.

\subsection{A cGAN with an Ensemble of Active Learning Discriminators (EAL-GAN)}
A cGAN generates labeled data by conducting an adversarial competition between two neural networks: a conditional generator G and a conditional discriminator D. The generator G, which takes the combination of class label information ${y^g}$ and a noise vector ${z \sim p(z)}$ as input, can directly generate realistic data ${x^g}$  of a given class ${y^g}$ to fool the discriminator. While the discriminator D is trained to: (i) distinguish the generated data ${{x^g} \sim {p^g}(x)}$  from the real data ${{x^r} \sim {p^r}(x)}$, and (ii) give the input data a probability distribution over its class labels. In such a framework, the generator is encouraged to synthesize more realistic samples conditioned on the class label ${y^g}$, so that the class probability predicted by the discriminator can be maximized. At the same time, the discriminator can be trained with more high-quality generated data. The two networks iteratively get their learning signal from the adversarial learning process until: (i) the generator’s outputs follow the data distribution conditioned on the given class label, and (ii) the discriminator can be an effective classifier for various classification tasks, for example, anomaly detection.

When applied to anomaly detection, a standard cGAN has two major limitations: (i) Training samples with class labels are very expensive to obtain. (ii) The class distributions of the training samples are highly imbalanced with far more normal samples than anomaly samples, this will lead to the discriminator heavily bias towards normal data thus resulting in a poor anomaly detector.

\subsubsection{Network Structure}
To overcome the limitations described above, we propose a conditional GAN with an ensemble of discriminators and an active learning strategy. The network structure of the proposed EAL-GAN is illustrated in Fig.1. To be more specific, EAL-GAN is composed of one generator G, and an ensemble of \emph{m} discriminators ${{{\mathop{\rm D}\nolimits} _1},{{\mathop{\rm D}\nolimits} _2}, \ldots ,{{\mathop{\rm D}\nolimits} _m}}$ which are trained based on an active learning algorithm.

\begin{figure*}[]
	\centering
	\includegraphics[scale=0.4]{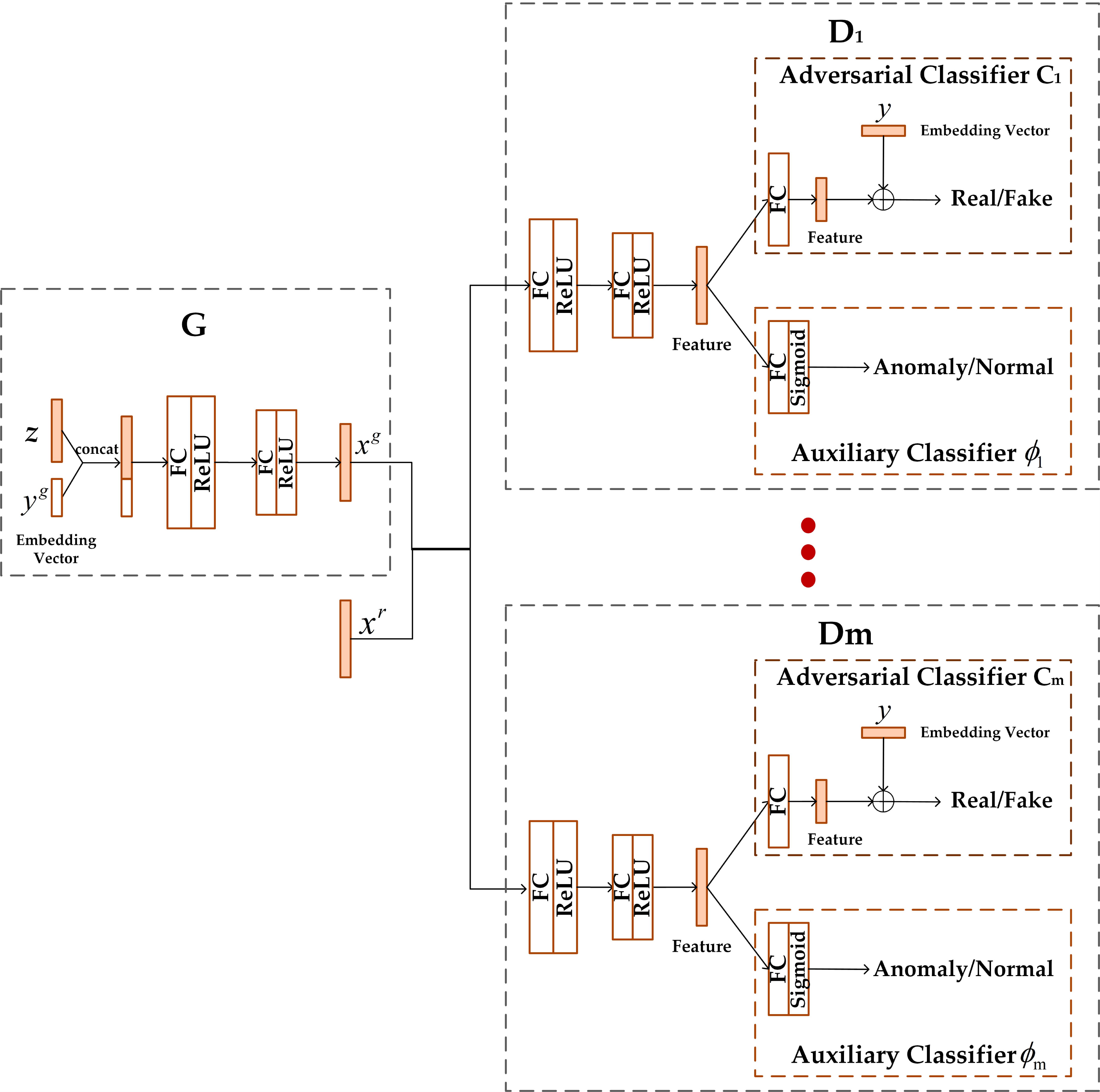}
	\caption{The network structures of the proposed EAL-GAN.G: Generator, ${{\mathop{\rm D}\nolimits} _i}$: the ${i}$-th discriminator (${1 \le i \le m}$), C: adversarial classifier in the discriminator that distinguishes the real data from fake, ${\phi}$: anomaly detector that predicts whether an input data is an anomaly. FC: Fully connected layer; ReLU: ReLU activation; ${ \oplus }$: inner product between two vectors.}
	\label{}
\end{figure*}

The generator G receives a random input vector ${z \sim p(z)}$ and an embedding vector of class label ${y^g}$ as its input, yielding ${x^g} = G(z,{y^g})$ as its output. This formulation allows G to generate data conditioned on label ${y^g}$, and the quality of G is evaluated by iteratively feeding the labeled data ${{x^g} = G(z,{y^g})}$ to the discriminators. Each of the discriminators is composed of two parts: (i) An adversarial classifier C which is used to distinguish the real data ${x^r}$ from the generated data ${x^g}$. (ii) An auxiliary classifier  aiming to classify the input of the discriminator into anomaly or normal classes. The output of ${\phi (x)}$ has a Sigmoid activation function so its outputs fall into the range [0, 1]. Anomaly detection is achieved through this auxiliary classifier, ${\phi (x)>0.5}$ indicates that ${x}$ is an anomaly, otherwise, ${x}$ is a normal data.

The output of the adversarial classifier C is the inner product between the embedded condition vector of class label ${y}$ and a feature vector extracted from the hidden layers of discriminator. The embedded condition vector will turn the adversarial classifier C into a projection discriminator \cite{RN50}, which incorporates the conditional information when distinguishing whether its input is real or fake.

Compared with other popular cGANs (e.g., AC-GAN \cite{RN51}), two essential improvements are made in the proposed EAL-GAN architecture: (i) Instead of putting only one conditional constraint on the generator (e.g., by adopting only one projection discriminator \cite{RN52}), in the proposed EAL-GAN, both the adversarial classifier C and the auxiliary classifier ${\phi}$ put conditional constraint on the generated data, such that the generated data can strictly follow the data distribution conditioned on the given class label. As we will show in Section 4.3, this dual-constraint strategy can help to stabilize the GAN training and improve the final performance. (ii) An ensemble of discriminators is used in EAL-GAN. Combined with the loss function in the following section, such ensemble architecture is expected to enhance the adversarial learning process and to better overcome the class imbalance issue.

\subsubsection{Ensemble Learning Loss Function}

One of the major problems in anomaly detection is that the data distribution is highly imbalanced with far more normal samples than anomalies. Even though EAL-GAN is a conditional GAN and we can generate class balanced fake samples to aid the design of the anomaly detector, the real data samples remain inherently highly imbalanced. EAL-GAN tackles this problem by introducing ensemble learning into its discriminator design. 

In an ideal case, the ensemble of discriminators should evaluate the generator from multi-perspective. However, this cannot be achieved by simply stacking several identical discriminators, as the generator can easily fool all the discriminators by simply generating identical fake data. To promote complementarity within the discriminators and rectify their bias towards the normal data, a new loss function is proposed. Given a batch of ${n_r}$ real data $\{ (x_i^r,y_i^r)\} _{i = 1}^{{n_r}}$  and a batch of ${n_f}$ fake data  $\{ (x_i^g,y_i^g)\} _{i = 1}^{{n_f}}$, the loss function of EAL-GAN has two parts: an adversarial loss ${{{\rm{{\cal L}}}_{GAN}}}$ and an auxiliary classifier loss ${{{\rm{{\cal L}}}_{AC}}}$. The adversarial loss for the \emph{k}-th discriminator (${1 \le k \le m}$) can be defined as follows:

\begin{footnotesize}
\begin{equation}
{\begin{array}{l}
{\rm{{\cal L}}}_{GAN}^{{D_k}}\! =\!  - \frac{1}{{{n_r}}}\sum\limits_{i = 1}^{{n_r}} {{{(1 - \overline {{p^k}} (y = 1|x_i^r))}^2}[\log C(x_i^r)]} \\
{\rm{        }} - \frac{1}{{{n_f}}}\sum\limits_{i = 1}^{{n_f}} {{{(1 - \overline {{p^k}} (y = 0|G({z_i},y_i^g)))}^2}[\log (1 - C(G({z_i},y_i^g)))]} 
\end{array}}
\end{equation}
\end{footnotesize}

${\rm{{\cal L}}}_{GAN}^{{D_k}}$ evaluates the discriminators' ability of distinguishing real and fake data and it consists of two parts: the loss on real data ${x^r}$ and loss on fake data ${x^g}$. To be more specific, two modulating factors ${{{(1 - \overline {{p^k}} (y = 1|x_i^r))}^2}}$ and ${{{(1 - \overline {{p^k}} (y = 0|G({z_i},y_i^g)))}^2}}$ have been added to the losses on the real data (i.e.,  ${- \log C(x_i^r)}$) and fake data (i.e.,${- \log (1 - C(G({z_i},y_i^g)))}$) respectively. In particular, ${\overline p (y = 1|x_i^r)}$ denotes the average probability that a real data ${x_i^r}$ being predicted as a real data by the previous \emph{k} adversarial classifiers, while ${\overline p (y = 0|G({z_i}|y_i^g))}$ denotes the average probability that a fake data $G({z_i}|y_i^g)$  being predicted as a fake data. The modulating factors assign higher weights to the data that have been wrongly classified by previous discriminators, therefore, forces the subsequent discriminators to focus on the deficiency of its previous members. By minimizing the loss in Eq.(2), the adversarial classifier ${{C_k}}$ of the \emph{k}-th discriminator can effectively determine whether its input data is a real data or generated by the generator.

The auxiliary classifier loss ${{{\rm{{\cal L}}}_{AC}}}$ is used to force the auxiliary classifier ${\phi}$ to effectively separate the anomaly and normal classes. The auxiliary classifier loss for the \emph{k}-th discriminator is defined as follows:

\begin{footnotesize}
	\begin{equation}
	{\begin{array}{l}
	{\rm{{\cal L}}}_{AC}^{{D_k}} \!= \! -\! \frac{1}{{{n_r}}}\sum\limits_{i = 1}^{{n_r}} {{{(1 - \overline {{p^k}} (y = y_i^r|x_i^r))}^2}[\log \phi (y = y_i^r|x_i^r)]} \\
	{\rm{   }} \!- \!\frac{1}{{{n_f}}}\sum\limits_{i = 1}^{{n_f}} {{{(1 - \overline {{p^k}} (y\! =\! y_i^g|G({z_i},y_i^g)))}^2}[\log \phi (y = y_i^g|G({z_i},y_i^g))]} 
	\end{array}}
	\end{equation}
\end{footnotesize}

Similarly, two modulating factors are added to the loss on the real and fake data, respectively. ${\overline {{p^k}} (y = y_i^r|x_i^r)}$ denotes the average probability assigned by previous \emph{k} auxiliary classifiers to the hypothesis that the real data ${x_i^r}$  belongs to its ground-true label ${y_i^r}$ , and ${\overline {{p^k}} (y = y_i^g|G({z_i},y_i^g))}$ denotes the average probability that the fake data ${G({z_i},y_i^g)}$ belongs to its label ${y_i^g}$.

The overall loss for the \emph{k}-th discriminator is:
\begin{equation}
{{\rm{{\cal L}}}_D^k = {\rm{{\cal L}}}_{GAN}^{{D_k}} + {\rm{{\cal L}}}_{AC}^{{D_k}}}
\end{equation}

The modulating factors in Eq.(2) and Eq.(3) always assign higher weights to the samples misclassified by previous discriminators. In an imbalanced dataset, these samples have higher chance to be the minority samples (anomalies) or the hardest samples for the previous discriminators. Minimizing a loss like this will force a discriminator adaptively focus on samples misclassified by the discriminators before it. In this way the discriminators work complementing each other. 

In the meantime, the loss for the generator consists of \emph{m} components, the \emph{k}-th of which is provided by the \emph{k}-th discriminator and can be divided into two parts: an adversarial loss ${{\rm{{\cal L}}}_{GAN}^{{G_k}}}$ and auxiliary classifier loss ${{\rm{{\cal L}}}_{AC}^{{G_k}}}$ , which are defined as follows:

\begin{footnotesize}
	\begin{equation}
	{{\rm{{\cal L}}}_{GAN}^{{G_k}} =  - \frac{1}{{{n_f}}}\sum\limits_{i = 1}^{{n_f}} {{{(1 - \overline {{p^k}} (y = 1|G({z_i},y_i^g)))}^2}[\log C(G({z_i},y_i^g))]} }
	\end{equation}
\end{footnotesize}
\begin{footnotesize}\vspace{1ex} 
	\begin{equation}
	{{\rm{{\!\cal L}\!}}_{AC}^{{D_k}}\! =\! -\! \frac{1}{{{n_f}}}\sum\limits_{i = 1}^{{n_f}} \!{{{(1 \!- \!\overline {{p^k}} (y = y_i^g|G({z_i},y_i^g)))}\!^2}[\log \phi (y\! =\! y_i^g\!|\!G({z_i}\!,\!y_i^g))]} }
	\end{equation}
\end{footnotesize}

The ensemble of discriminators evaluates the generator in a sequential manner. When the \emph{k}-th discriminator evaluates the generator, the modulating factors in Eq.(5) and Eq.(6) assign higher weights to the fake data that previous discriminators have less confidence on. In such a sequential evaluation process, the subsequent discriminators always concentrate on the deficiencies of its previous members, and the generator can be evaluated from diverse and complementary perspectives. 

By summing the losses provided by all \emph{m} discriminators, we can obtain the overall loss for the generator as follows:

\begin{equation}
{{{\rm{{\cal L}}}_G} = \sum\limits_{k = 1}^m {({\rm{{\cal L}}}_{GAN}^{{G_k}} + } {\rm{{\cal L}}}_{AC}^{{G_k}})}
\end{equation}

By iteratively updating the discriminators and generators with loss functions in Eq.(4) and Eq.(7), the EAL-GAN maintains an intensive competition within its framework until the whole system converges.

\subsubsection{Active Learning}
Another major difficulty in anomaly detection is that it can be very expensive to label sufficiently large number of training samples. EAL-GAN approaches this issue from two perspectives. \emph{First}, it uses conditional GAN to generate class balanced fake data to aid the training of the auxiliary classifier ${\phi}$. Class balance in the generated data can be achieved by controlling the condition ${y^g}$ in the generator. \emph{Second}, EAL-GAN adopts an active learning strategy to reduce the burden on labeling real training samples.

In EAL-GAN, the auxiliary classifier ${\phi}$ in each discriminator outputs its classification confidence on the hypothesis that the input data is an anomaly, this can be naturally used to develop an active learning strategy. Specifically, given a batch of ${{n_{bs}}}$ unlabeled real samples, EAL-GAN first feeds all these samples to each of the discriminators, and then obtains an anomaly score $\overline {\phi (x)} $  by averaging the predictions of all the auxiliary classifiers ${\phi}$. Only samples that have a score closest to 0.5 will be labeled (anomaly or normal), thus greatly reducing the number of samples need to be labeled. These samples are chosen because: (i) They are located near the classification boundary and carry the most information, (ii) according to the results in margin-based methods (i.e., Support Vector Machine and Boosting), these samples account for only a small proportion of the training data, thus require low annotation cost. (iii) In a highly imbalanced distribution, these samples have higher chance to be anomalies, therefore can largely alleviate the class imbalance issue in anomaly detection. The selected real data and a batch of fake data generated by the generator will be used to update all the discriminators. In EAL-GAN, a hyper-parameter $\rho $ is used to determine the proportion of data selected by the active sampling.

\subsubsection{Anomaly Detection}
Once the training of EAL-GAN has converged, the auxiliary classifier ${\phi}$ of all the discriminators can form a powerful ensemble that can effectively classify anomaly and normal data. Specifically, when determining whether an instance ${x}$ is an anomaly, EAL-GAN feeds ${x}$ to all the discriminators, and then obtain the anomaly score $\overline {\phi (x)} $ by averaging the outputs of all the auxiliary classifier ${\phi}$. If $\overline {\phi (x)}  > 0.5$ , then ${x}$ is predicted as an anomaly, otherwise, ${x}$ is a normal data. The entire EAL-GAN anomaly detection algorithm is described in Algorithm 1.

\begin{algorithm}[!ht]
	\caption{EAL-GAN}
	\label{alg_1}
	
	\begin{algorithmic}
		\REQUIRE
			${x_{unlabeld}^r}$:unlabeled real data; ${z \sim {p^g}(z)}$:noise distribution; ${{y^g} \sim {p^g}(y)}$:label for the conditional generator; ${n_{bs}}$:mini-batch size; $\rho $:the ratio of active sampling; \emph{m}:the number of discriminators \\
	
	\end{algorithmic}
	\hrulefill
	\hspace*{0.02in}{\bf Train:}
	\begin{algorithmic}[1]
		\FOR {each training epoch}
			\FOR {each training iteration}
				\STATE Generating a batch of ${n_{bs}}$ fake data ${x^g}$  with the generator from random noise ${z}$ and given label  ${y^g}$.
				
				\STATE Use ${x^g}$ , ${y^g}$ to update the generator by minimizing the loss in Eq.(7)
				
				\STATE Given a batch of ${n_{bs}}$ unlabeled real samples ${x_{unlabeld}^r}$, use the active sampling strategy described in Section 3.2.3 to select ${n_r} = {n_{bs}} \times \rho $ samples from ${x_{unlabeld}^r}$ (denoted as ${x^r}$) and annotate their class labels ${y^r}$.
				
				\STATE Use ${x^r}$, ${y^r}$, ${x^g}$, ${y^g}$  to calculate the adversarial loss in Eq.(2) and auxiliary classifier loss in Eq.(3) for all the discriminators
				
				\STATE Sequentially update all the discriminators ${{\mathop{\rm D}\nolimits} _1},{{\mathop{\rm D}\nolimits} _2},...,{{\mathop{\rm D}\nolimits} _m}$ by minimizing the loss in Eq.(4)
				
			\ENDFOR
		\ENDFOR
		
	\end{algorithmic}
	\hrulefill
	
    \hspace*{0.02in}{\bf Anomaly Detection}
    \begin{algorithmic}[1]
    	\STATE Feed an unknown instance \emph{x} to all the discriminators, then calculate the average anomaly score $\overline {\phi (x)} $ by averaging the outputs of all the auxiliary classifier $\phi (x)$. If $\overline {\phi (x)}  > 0.5$ then \emph{x} is predicted as an anomaly, otherwise, \emph{x} is a normal data
    \end{algorithmic}

\end{algorithm}

\subsection{Advantages of the EAL-GAN Anomaly Detection Method}
There are two insights behind the design of EAL-GAN. \emph{First}, previous studies have shown that forcing a neural network to perform additional tasks is known to improve performance on the original task \cite{RN51},\cite{RN53}. Therefore, introducing an appropriate multi-task structure into the GAN framework should result in better performance. \emph{Second}, GAN training can be unstable and one commonly accepted reason for its instability is that gradients passing from the discriminator to the generator become uninformative when the discriminator focus too much on the global structure or local details \cite{RN54}. Evaluating generator from multi-perspective can stabilize the GAN training. These insights motivate us to adopt multiple multi-task discriminators in the EAL-GAN. As a result, EAL-GAN has some distinctive advantages over most existing anomaly detectors: 

\emph{First}, the cost of annotating anomaly data for training a fully supervised anomaly detector is significantly reduced. The class imbalance issue is one of the major challenges in constructing an effective anomaly detector. In a dataset where only few anomalies are collected and labeled, the anomalies can easily be misclassified by a standard classifier. The loss function in Eq.(4) forces each of the discriminators to adaptively concentrate more on these instances. Such a strategy of overcoming class imbalance can not only allow the discriminators to effectively learn from the real data, it also provides better learning signal to the generator. And in turn, the generator can effectively generate sufficiently balanced and high-quality fake data for training the discriminators. Moreover, the proposed active sampling strategy constantly selects data carrying the most information but account for relatively small proportion of the available real data. Training the discriminators with the combination of limited number of real data and unlimited number of generated data can greatly reduce the requirements for labeled anomalies.

\begin{figure*}[!t]
	\centering
	\subfloat[]{\includegraphics[width=2.3in, height=1.9in]{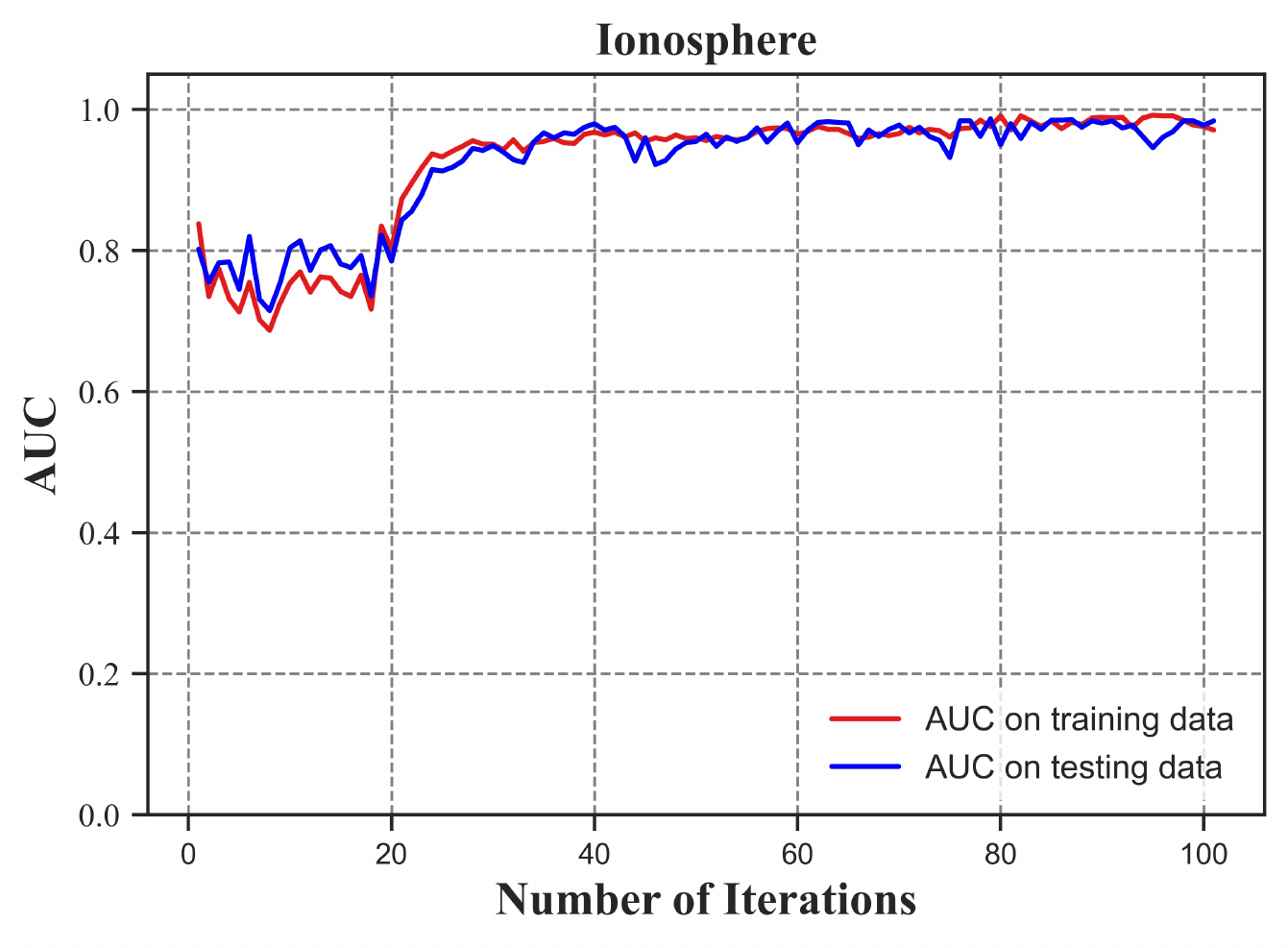}
		\label{fig_2a_case}}
	\hfil
	\subfloat[]{\includegraphics[width=2.3in, height=1.9in]{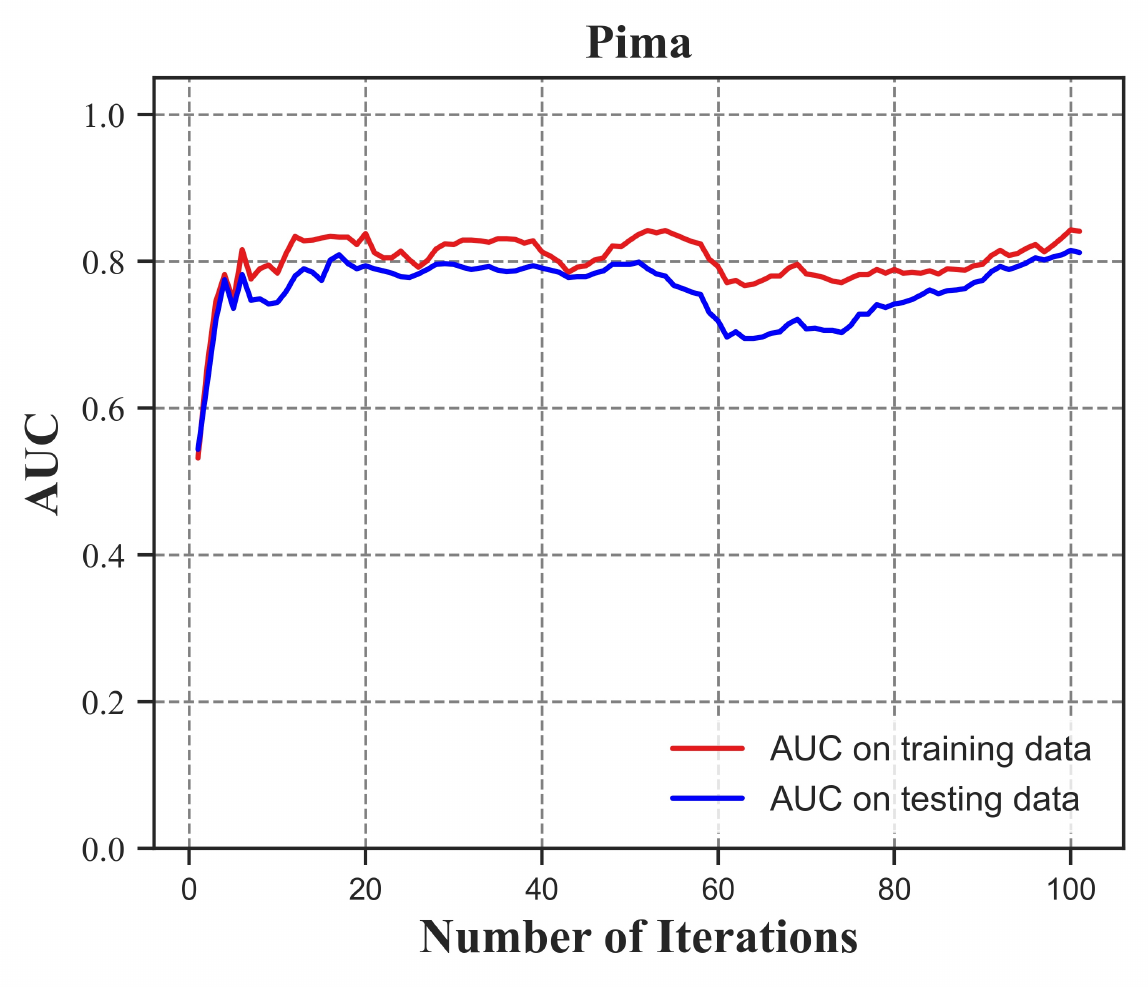}
		\label{fig_2b_case}}
	\hfil
	\subfloat[]{\includegraphics[width=2.3in, height=1.9in]{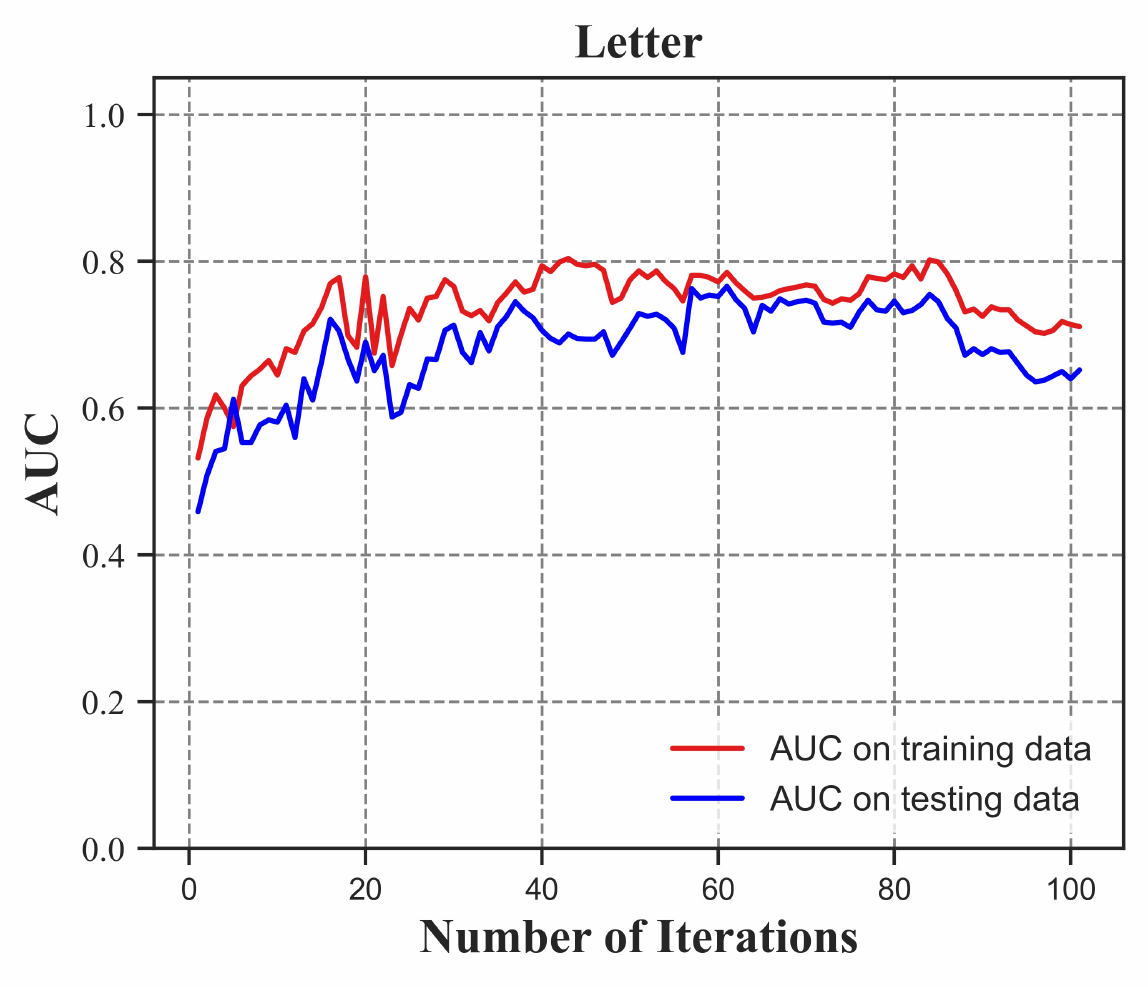}
		\label{fig_2c_case}}
	
	\subfloat[]{\includegraphics[width=2.3in, height=1.9in]{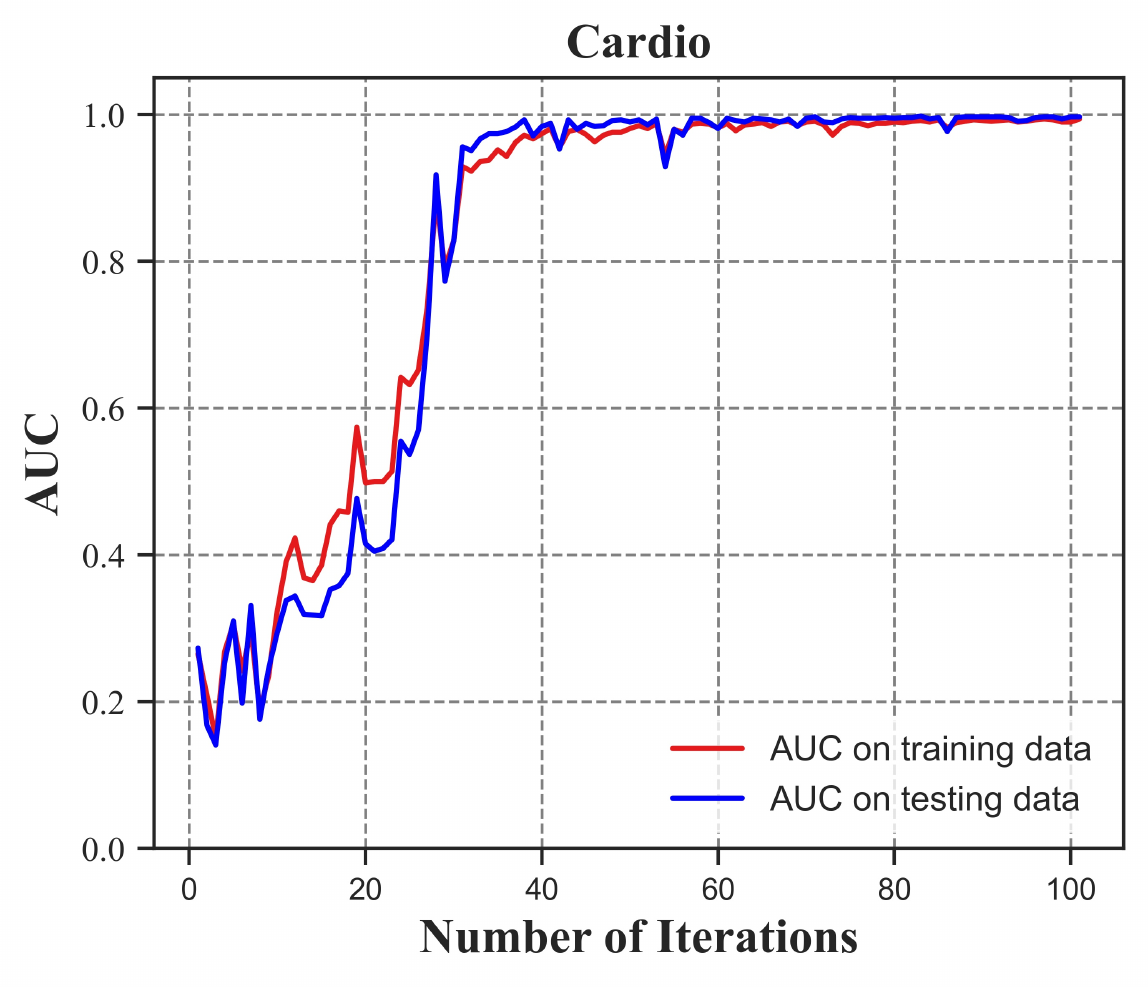}
		\label{fig_2d_case}}
	\hfil
	\subfloat[]{\includegraphics[width=2.3in, height=1.9in]{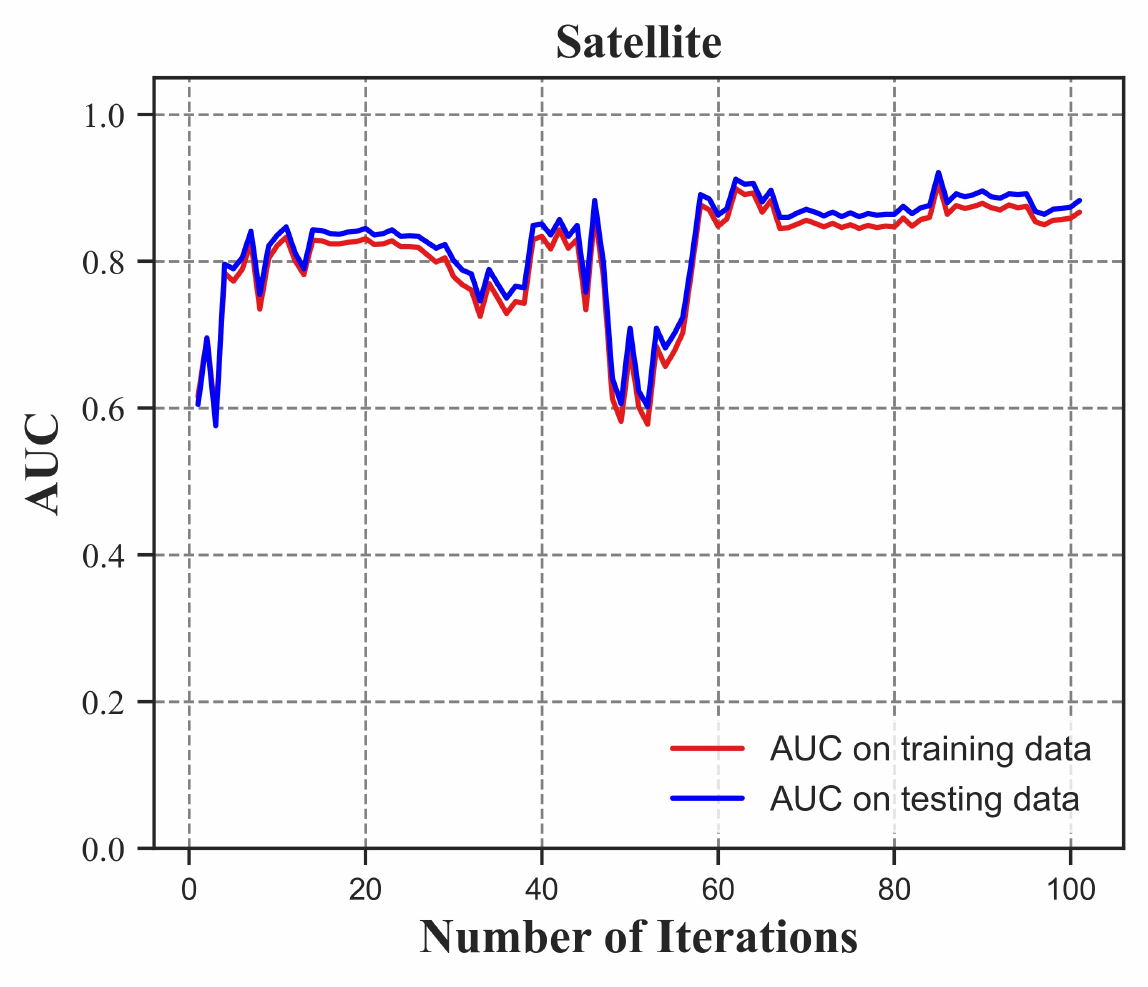}
		\label{fig_2e_case}}
	\hfil
	\subfloat[]{\includegraphics[width=2.3in, height=1.9in]{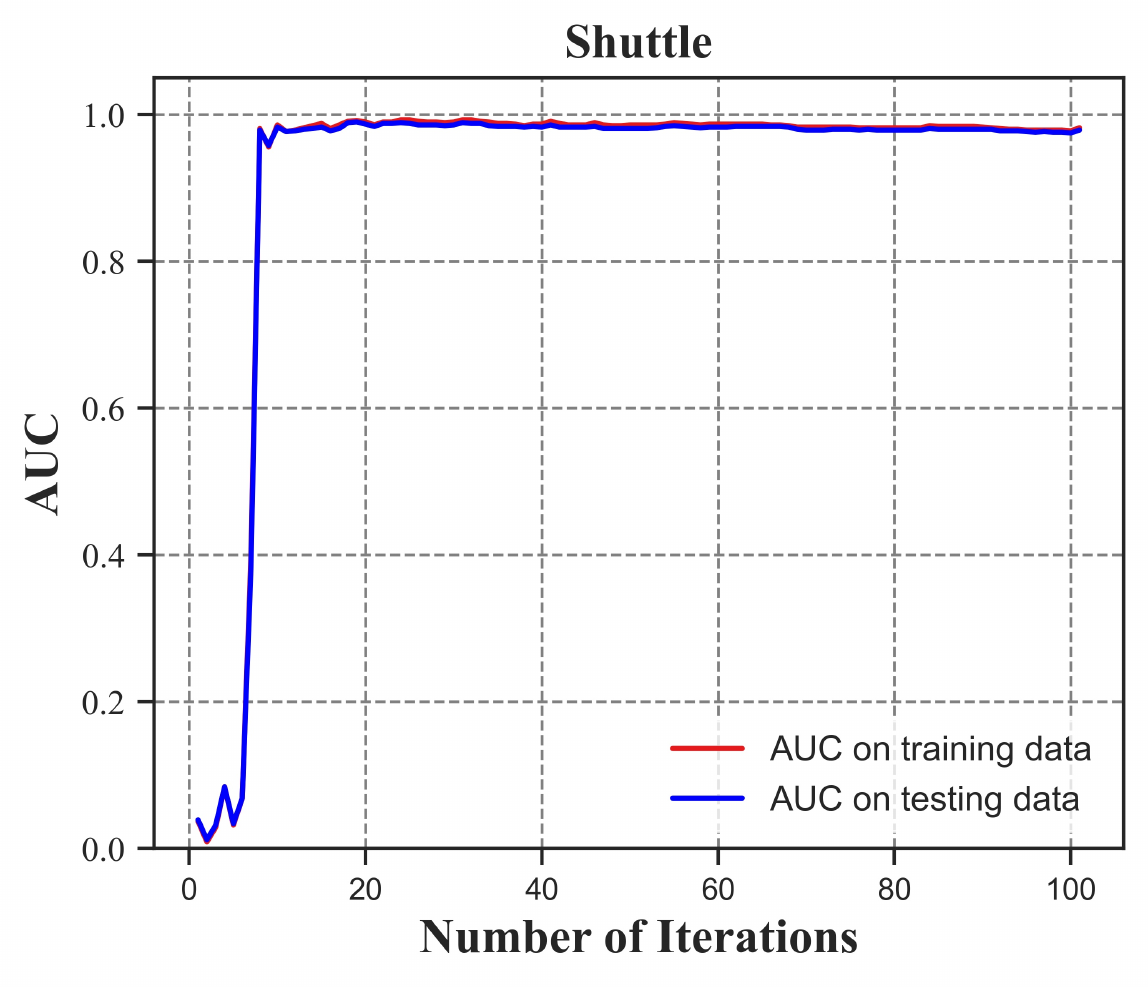}
		\label{fig_2f_case}}
	
	\caption{EAL-GAN’s performance on six datasets. The red line denotes the AUC on training set, while blue denotes the AUC on testing set.}
	\label{fig_2}
\end{figure*}

\emph{Second}, strong regularizations are introduced into the training process. Regularization is a common technique of preventing a neural network from overfitting, and this is often achieved by turning the model training from a single-objective optimization to a multi-objective optimization. In EAL-GAN, the discriminator is forced to jointly perform three tasks: (i) distinguishing the real data from the fake data, (ii) predicting if its input data is an anomaly, (iii) competing with the generator. When performing the first task, an extra embedding layer has been introduced into the specific branch (please see the adversarial classifier C in Fig.1). This can turn the discriminator architecture into a projection discriminator. Miyato and Koyama \cite{RN50} have theoretically proven that such a modification can bring extra regularization into cGAN training. Moreover, the unique architecture of multiple discriminators and its corresponding loss function can inject stronger competition into the GAN training, which in turn forces the discriminators to be better models on the specific tasks. Such a multi-task structure and the extra regularizations allow the EAL-GAN to show greater resistance to overfitting. To better illustrate this merit, we plot the Area Under Receiver Operating Characteristic Curve (AUC) of EAL-GAN during the training process on six datasets (i.e., \emph{Ionosphere}, \emph{Pima}, \emph{Letter}, \emph{Cardio}, \emph{Satellite} and \emph{Shuttle}, whose detailed information can be found in the following section). Each dataset is randomly divided into a training set (60\%) and a testing set (40\%). The AUCs of the training and testing sets are illustrated in Fig.2. Despite the variety in sample size, feature dimension and anomaly ratio, EAL-GAN shows consistent performance on training and testing set. This can be an attractive property for practitioners, as by choosing the model with best performance on training set, one can expect to obtain anomaly detector with excellent generalization.

\section{Experiment}
\subsection{Experimental settings}
In this section, we conduct extensive experiments to evaluate the proposed EAL-GAN, the robustness of hyper-parameter setting, and the contribution of each component in EAL-GAN (ablation study). The datasets, performance measures, competing methods and parameter setting involved in the experiments are described as follows:

\subsubsection{Datasets}
We conduct experiments on both real and synthetic datasets. While the real datasets are adopted to demonstrate the effectiveness of EAL-GAN in real-world applications, synthetic datasets of various distributions are created to reveal the contribution of each essential improvement made in EAL-GAN. In particular, 20 real-world datasets which are frequently adopted as the benchmarks in this area have been used. These datasets cover a wide variety in sample size, feature dimension, and anomaly ratio. Their detailed information is summarized in Table 1, ordered by their increasing sample size.

\begin{table}[htbp]
	\centering
	\caption{Detailed Information of the Real-World Datasets
		(\#Abbr: Abbreviation of Datasets, \#R: Anomaly Ratio, \#D: Feature Dimension, \#Inst: Number of Training data)}
	\label{Table_1}
	\begin{tabular}{lcccc}
		\toprule[1.5pt]
		\multicolumn{1}{l}{Dataset} & \#Abbr & \#Inst & \#D   & \#R \\
		\midrule[1.5pt]
		Lympho & Lym   & 148   & 18    & 4.05\% \\
		Glass & Gla   & 214   & 9     & 4.21\% \\
		Ionosphere & Ion   & 351   & 33    & 35.90\% \\
		Arrhythmia & Arr   & 452   & 274   & 14.60\% \\
		Pima  & Pim   & 768   & 8     & 34.90\% \\
		Vowels & Vow   & 1456  & 12    & 3.43\% \\
		Letter & Let   & 1600  & 32    & 6.25\% \\
		Cardio & Cad   & 1831  & 21    & 9.61\% \\
		Musk  & Mus   & 3062  & 166   & 3.17\% \\
		Optdigits & Opt   & 5216  & 64    & 2.88\% \\
		Satimage2 & Sat2  & 5803  & 36    & 1.22\% \\
		Satellite & Sat   & 6435  & 36    & 31.64\% \\
		Pendigits & Pen   & 6870  & 16    & 2.27\% \\
		Annthyroid & Ann   & 7200  & 21    & 7.42\% \\
		Mnist & Mni   & 7603  & 100   & 9.21\% \\
		Campaign & Cam   & 41188 & 62    & 11.27\% \\
		Shuttle & Shu   & 49097 & 9     & 7.15\% \\
		Celeba & Cel   & 202599 & 39    & 2.24\% \\
		Fraud & Fra   & 284807 & 29    & 0.17\% \\
		Donors & Don   & 619326 & 10    & 5.93\% \\
		\bottomrule[1.5pt]
	\end{tabular}%
\end{table}%

As for the synthetic datasets, we vary four different factors (i.e., cluster-type, feature dimension, anomaly ratio and sample size) to create various datasets. Cluster type is used to investigate the performance of EAL-GAN on different data distributions. A group of datasets with different cluster types (e.g., single-cluster, multi-cluster and multi-density) are created and Fig.3 illustrates the distribution. We vary the sample size (from 1000 to 20000), feature dimension (from 2 to 160) and anomaly ratio (from 2\% to 20\%) to create 20 datasets for each given cluster type. Following the setting in [17], all the anomalies are randomly sampled from a uniform distribution. Detailed information of the synthetic data is given in Table 2.

\begin{table}[htbp]
	\centering
	\caption{Description of the Synthetic Datasets}
	\begin{tabular}{cccc}
		\toprule[1.5pt]
		Cluster type & \#Inst & \#D   & \#R \\
		\midrule[1.5pt]
		Single-cluster & 1000-20000 & 2-160 & 2\%-20\% \\
		Multi-cluster & 1000-20000 & 2-160 & 2\%-20\% \\
		Multi-density & 1000-20000 & 2-160 & 2\%-20\% \\
		\bottomrule[1.5pt]
	\end{tabular}%
	\label{tab:addlabel}%
\end{table}%

\begin{figure}[H]
	\centering
	\subfloat[Single-cluster]{\includegraphics[width=1in, height=1in]{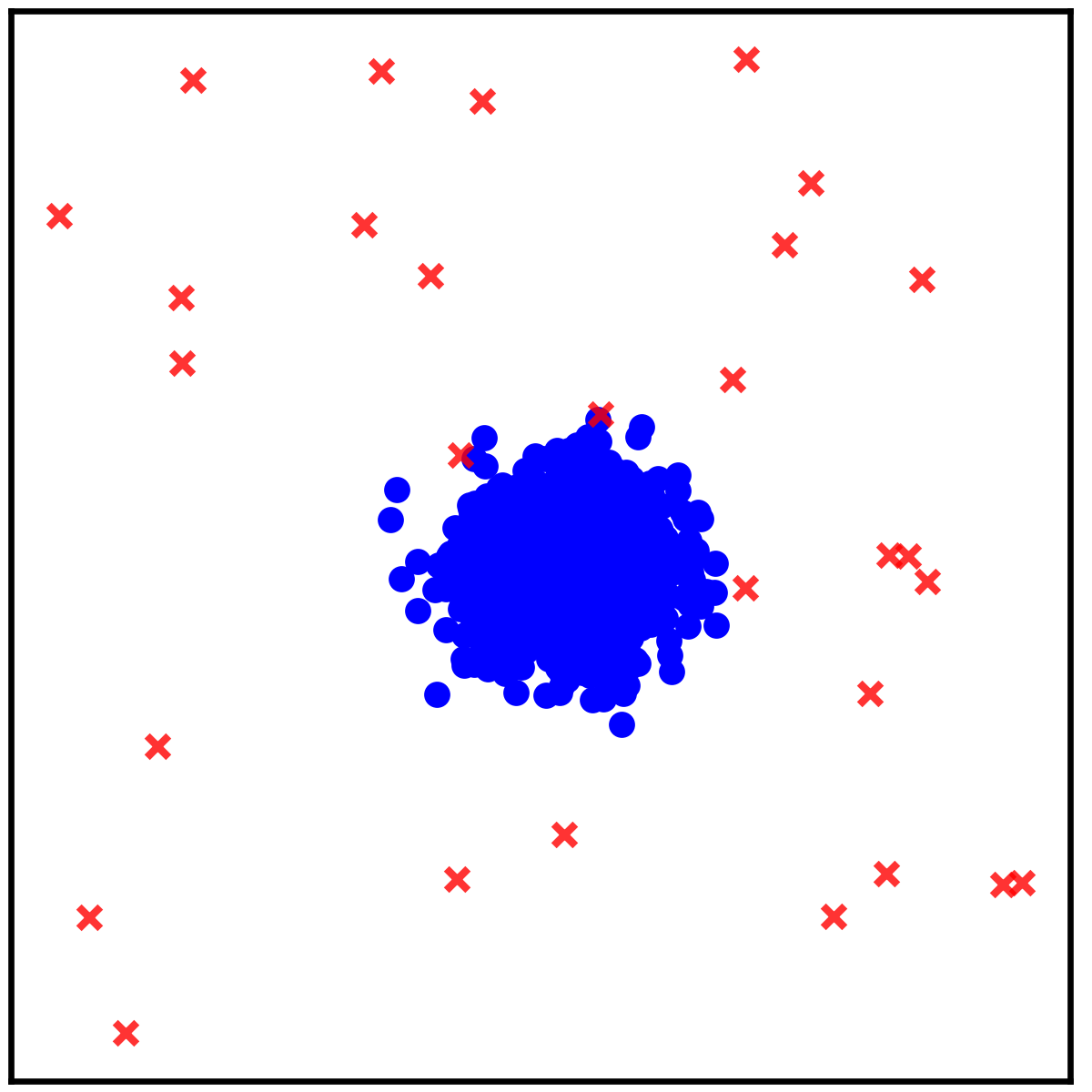}
		\label{fig_3a_case}}
	\hfil
	\subfloat[Multi-cluster]{\includegraphics[width=1in, height=1in]{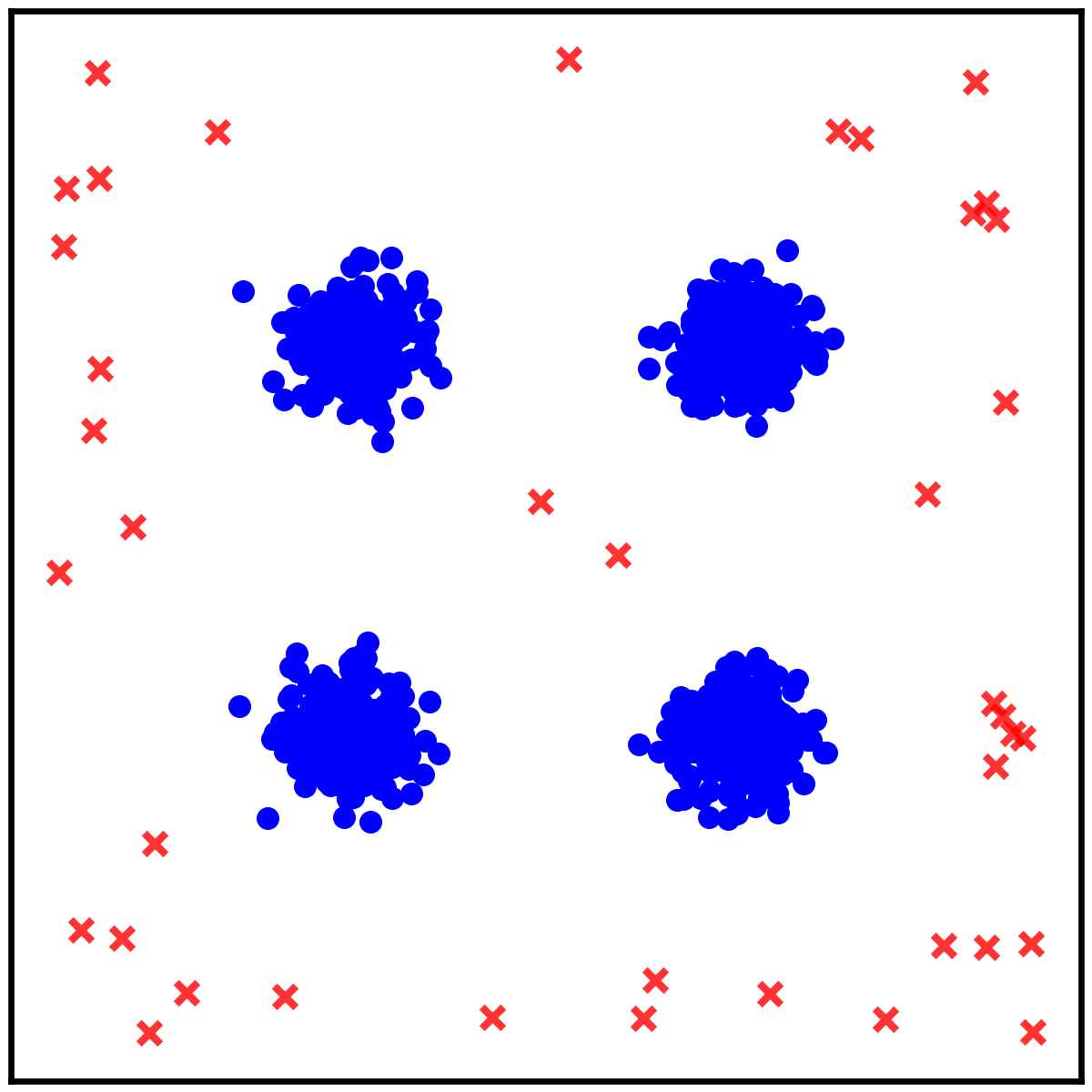}
		\label{fig_3b_case}}
	\hfil
	\subfloat[Multi-density]{\includegraphics[width=1in, height=1in]{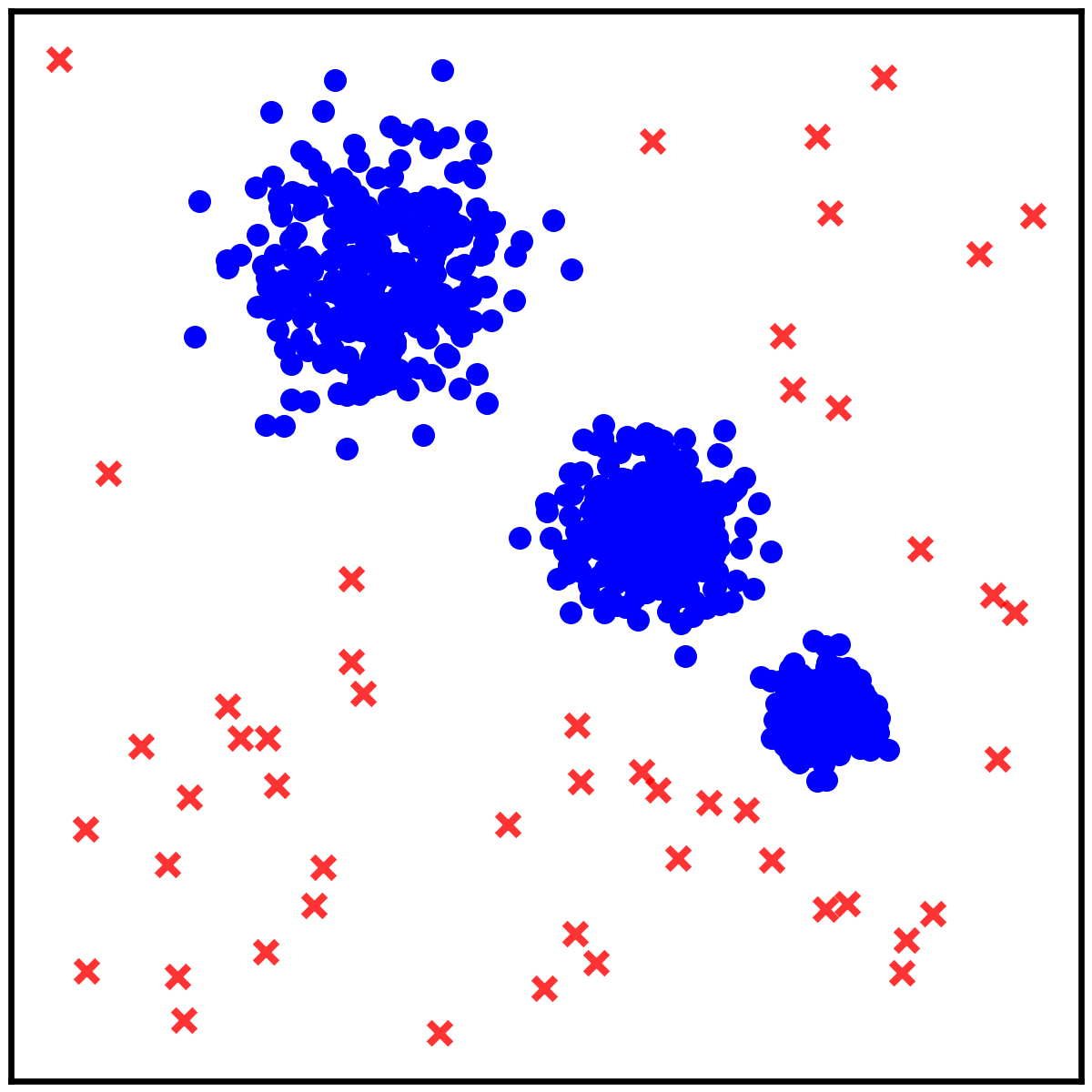}
		\label{fig_3c_case}}
	
	\caption{Illustration of three cluster types. The blue dot symbol represents normal data, and the red \emph{X} symbol denotes the anomalies.}
	\label{fig_3}
\end{figure}

\subsubsection{Competing methods and parameter settings}
EAL-GAN is compared with 9 competing methods, and these methods can be grouped into 6 categories: (i) one density-based methods, LOF \cite{RN55}, (ii) two deep anomaly detection methods, DSVDD \cite{RN20} and deviation network (DevNet) \cite{RN8}, (iii) one cluster-based method, CBLOF \cite{RN56}, (iv) two ensemble detectors, iForest \cite{RN36} and Feature Bagging (FB) \cite{RN57}, (v) two distance-based methods, AvgKNN \cite{RN58} and FastABOD (FOD) \cite{RN59}, (vi) one GAN-based method, MOGAAL \cite{RN17}. These methods are chosen because they are the state-of-the-art in relevant areas, i.e., DSVDD in the deep feature learning for anomaly detection, DevNet in deep anomaly detection with limited labeled data, iForest and FB in ensemble detectors, MOGAAL \cite{RN17} in GAN-based methods. The codes of some methods are extracted from PyOD \cite{RN60}, while the codes of the proposed EAL-GAN are openly available on Github\footnote{https://github.com/smallcube/EAL-GAN}.

We have tried our best to set optimal parameters for all the methods. For AvgKNN, LOF and CBLOF, since their performance can be significantly affected by the number of nearest neighborhoods considered, we search the best number in the range of ${[2,4,6, \ldots ,\sqrt {\frac{n}{{100}}} ]}$, where \emph{n} is the size of training set. To build optimal deep anomaly detectors, we largely follow the parameter settings suggested by \cite{RN8}: (i) a three-hidden-layer network architecture is built for both DSVDD and DevNet. Specifically, 1000 neural units are built in the first hidden layer, followed by 250 and 20 units in the second and third layers, respectively. (ii) Both DevNet and DSVDD take the Root Mean Square propagation (RMSprop) optimizer to perform gradient descent optimization and the maximum training epoch is set to 50, with 20 mini-batches in each epoch. To ensure a fair comparison, all the ensemble detectors, i.e., iForest, FB, MOGAAL and the proposed EAL-GAN, build 10 ensemble members in their architectures.

As for the proposed EAL-GAN, the parameter setting can be described as follows: (\textbf{i}) one generator and 10 discriminators are built for each dataset, (\textbf{ii}) a four-layer neural network (128*2\emph{d}*2\emph{d}*\emph{d}) is built for the generator, and a three-layer neural network(\emph{d}*2\emph{d}*(1+1)) is built for each discriminator. While the generator takes a latent vector of size 128 as input, \emph{d} represents the feature dimension of the input data. The anomaly detector (i.e., the auxiliary classifier ${\phi}$ in discriminator) and the adversarial classifier share the first two layers of the discriminator neural network (please see Fig.1). (\textbf{iii}) The generator and discriminator use Adam optimizer to perform gradient descent, and the learning rate of the generator is set as 0.01. The learning rate for each discriminator is a random value in the range [0.01, 0.05]. Training each discriminator with different learning rate can further promote the diversity within the ensemble. (\textbf{iv}) The mini-batch size is set as 128. (\textbf{v}) The sampling ratio is set as 5\%, indicating that only 5\% of the real data in each mini-batch will be selected and labeled. (\textbf{vi}) The maximum training epoch is set as 50. During the training phase, we record the model with the best empirical performance (measured by ${\mathop{\rm AUC}\nolimits}  \times Gmean$ ) on training set as the final anomaly detector, and report its performance on testing set. More information can be found in our codes.

\subsubsection{Performance measures and statistical tests}
We use two popular performance measures, the AUC and ${Gmean}$ to evaluate all the anomaly detectors. AUC summarizes the Receiver Operating Characteristic Curve (ROC) of true positives against false positives. An AUC value of 1 indicates the best performance, while an AUC value of 0.5 indicates that the corresponding model makes random predictions. ${Gmean}$ is defined as follows:

\begin{equation}
{Gmean = \sqrt {T{P_{rate}} \times T{N_{rate}}} }
\end{equation}

Where ${T{P_{rate}}}$ denotes the ratio of positive examples (anomalies) that have been correctly predicted as positive, and ${T{N_{rate}}}$ represents the ratio of negative example (normal data) that have been correctly predicted as negative. ${Gmean}$ is affected by the absolute detection accuracy on both anomaly and normal data. The reported AUC and ${Gmean}$ are averaged results over 10 independent runs, and in each run, 60\% of the available data are randomly sampled as the training data and the remaining 40\% data are used as the testing data. In addition, we perform non-parametric statistical tests, i.e, Friedman test \cite{RN61} and its post-hoc test (Nemenyi test \cite{RN62}), on the obtained performance measures. The null hypothesis of Friedman test is that there is no significant difference in all the involved methods. If the statistical test result exceeds the critical value, the null hypothesis will be rejected and the post-hoc test will be conducted to identify the best method.

\subsection{Experimental Results on Real-world Datasets}
In this sub-section, we evaluate the proposed EAL-GAN against 9 other competitive anomaly detectors on 20 real-world datasets. Table 3 and 4 present the performance of all the anomaly detectors in term of AUC and ${Gmean}$, respectively. The last rows in Table 3 and 4 give the average rankings obtained from the Friedman test. The best result for each dataset is highlighted in red color.

\begin{table*}[htbp]
	\centering
	\caption{Average AUCs obtained by all the involved anomaly detectors on the real-world datasets.}
	\begin{tabular}{ccccccccccc}
		\toprule[1.5pt]
		Data  & LOF   & DSVDD & DevNet & CBLOF & iForest & FB    & AvgKNN & FOD   & MOGAAL & EAL-GAN \\
		\midrule[1.5pt]
		Lym   & 0.977 & 0.976 & \color{red}{\textbf{1.000}} & 0.963 & 0.993 & 0.977 & 0.974 & 0.911 & 0.910  & 0.951 \\
		Gla   & 0.864 & 0.632 & 0.987 & 0.862 & 0.746 & 0.851 & 0.851 & 0.795 & 0.529 & \color{red}{\textbf{0.977}} \\
		Ion   & 0.875 & 0.842 & 0.894 & 0.899 & 0.851 & 0.872 & 0.927 & 0.925 & 0.874 & \textcolor[rgb]{1.000, 0.000, 0.000}{\textbf{0.969}} \\
		Arr   & 0.779 & 0.781 & 0.749 & 0.778 & \textcolor[rgb]{1.000, 0.000, 0.000}{\textbf{0.820}} & 0.778 & 0.786 & 0.769 & 0.751 & \textcolor[rgb]{1.000, 0.000, 0.000}{\textbf{0.820}} \\
		Pim   & 0.627 & 0.622 & 0.656 & 0.656 & 0.676 & 0.624 & 0.708 & 0.679 & 0.758 & \textcolor[rgb]{1.000, 0.000, 0.000}{\textbf{0.825}} \\
		Vow   & 0.941 & 0.780  & \textcolor[rgb]{1.000, 0.000, 0.000}{\textbf{0.999}} & 0.920  & 0.751 & 0.940  & 0.968 & 0.961 & 0.803 & 0.982 \\
		Let   & 0.859 & 0.612 & 0.848 & 0.782 & 0.623 & 0.864 & 0.877 & \textcolor[rgb]{1.000, 0.000, 0.000}{\textbf{0.878}} & 0.669 & 0.832 \\
		Cad   & 0.574 & 0.935 & 0.977 & 0.807 & 0.924 & 0.598 & 0.724 & 0.569 & 0.914 & \textcolor[rgb]{1.000, 0.000, 0.000}{\textbf{0.997}} \\
		Mus   & 0.529 & \textcolor[rgb]{1.000, 0.000, 0.000}{\textbf{1.000}} & 0.968 & \textcolor[rgb]{1.000, 0.000, 0.000}{\textbf{1.000}} & 0.999 & 0.520  & 0.799 & 0.816 & 0.798 & \textcolor[rgb]{1.000, 0.000, 0.000}{\textbf{1.000}} \\
		Opt   & 0.550  & 0.500   & \textcolor[rgb]{1.000, 0.000, 0.000}{\textbf{1.000}} & 0.771 & 0.699 & 0.553 & 0.629 & 0.533 & 0.690  & \textcolor[rgb]{1.000, 0.000, 0.000}{\textbf{1.000}} \\
		Sat2  & 0.542 & 0.998 & 0.963 & \textcolor[rgb]{1.000, 0.000, 0.000}{\textbf{0.999}} & 0.996 & 0.538 & 0.954 & 0.819 & 0.961 & 0.993 \\
		Sat   & 0.557 & 0.662 & 0.773 & 0.749 & 0.701 & 0.557 & 0.684 & 0.571 & 0.678 & \textcolor[rgb]{1.000, 0.000, 0.000}{\textbf{0.899}} \\
		Pen   & 0.530  & 0.930  & \textcolor[rgb]{1.000, 0.000, 0.000}{\textbf{0.998}} & 0.885 & 0.950  & 0.533 & 0.749 & 0.688 & 0.976 & \textcolor[rgb]{1.000, 0.000, 0.000}{\textbf{0.998}} \\
		Ann   & 0.673 & 0.749 & 0.773 & 0.606 & 0.632 & 0.74  & 0.693 & 0.730  & 0.690  & \textcolor[rgb]{1.000, 0.000, 0.000}{\textbf{0.949}} \\
		Mni   & 0.716 & 0.853 & 0.694 & 0.849 & 0.801 & 0.719 & 0.848 & 0.782 & 0.909 & \textcolor[rgb]{1.000, 0.000, 0.000}{\textbf{0.996}} \\
		Cam   & 0.628 & 0.748 & 0.819 & 0.737 & 0.707 & 0.615 & 0.74  & 0.740  & 0.780  & \textcolor[rgb]{1.000, 0.000, 0.000}{\textbf{0.921}} \\
		Shu   & 0.526 & 0.992 & 0.978 & 0.612 & \textcolor[rgb]{1.000, 0.000, 0.000}{\textbf{0.997}} & 0.518 & 0.654 & 0.623 & 0.907 & 0.992 \\
		Cel   & 0.546 & 0.944 & 0.950  & 0.763 & 0.687 & 0.521 & 0.605 & 0.723 & 0.758 & \textcolor[rgb]{1.000, 0.000, 0.000}{\textbf{0.959}} \\
		Fra   & 0.512 & \textcolor[rgb]{1.000, 0.000, 0.000}{\textbf{0.977}} & 0.964 & 0.952 & 0.952 & 0.528 & 0.938 & 0.856 & 0.955 & 0.969 \\
		Don   & 0.581 & 0.955 & \textcolor[rgb]{1.000, 0.000, 0.000}{\textbf{1.000}} & 0.754 & 0.772 & 0.627 & 0.636 & 0.653 & 0.830  & \textcolor[rgb]{1.000, 0.000, 0.000}{\textbf{1.000}} \\
		Avg   & 0.669 & 0.824 & 0.900   & 0.817 & 0.814 & 0.674 & 0.787 & 0.751 & 0.807 & 0.951 \\
		Rank  & 7.65  & 5.28  & 3.55  & 5.38  & 5.45  & 7.75  & 5.45  & 6.53  & 5.85  & 2.13 \\
		\bottomrule[1.5pt]
	\end{tabular}%
	\label{table_3}%
	
\end{table*}

\begin{table*}[htbp]
	\centering
	\caption{Average ${Gmean}$ obtained by all the involved anomaly detectors on the real-world datasets}
	\begin{tabular}{ccccccccccc}
		\toprule[1.5pt]
		Data  & LOF   & DSVDD & DevNet & CBLOF & iForest & FB    & AvgKNN & FOD   & MOGAAL & EAL-GAN \\
		\midrule[1.5pt]
		Lym   & 0.813 & 0.813 & 0.728 & 0.813 & \textcolor[rgb]{1.000, 0.000, 0.000}{\textbf{0.857}} & 0.813 & 0.813 & 0.590  & 0.801 & 0.721 \\
		Gla   & 0.233 & 0.283 & 0.560  & 0.142 & 0.142 & 0.291 & 0.142 & 0.285 & 0.187 & \textcolor[rgb]{1.000, 0.000, 0.000}{\textbf{0.777}} \\
		Ion   & 0.772 & 0.767 & 0.596 & 0.831 & 0.722 & 0.773 & 0.892 & 0.879 & 0.634 & \textcolor[rgb]{1.000, 0.000, 0.000}{\textbf{0.901}} \\
		Arr   & 0.624 & 0.645 & 0.503 & 0.625 & \textcolor[rgb]{1.000, 0.000, 0.000}{\textbf{0.674}} & 0.610  & 0.634 & 0.582 & 0.645 & 0.673 \\
		Pim   & 0.569 & 0.582 & 0.400   & 0.593 & 0.616 & 0.568 & 0.640  & 0.622 & 0.330  & \textcolor[rgb]{1.000, 0.000, 0.000}{\textbf{0.721}} \\
		Vow   & 0.583 & 0.519 & 0.646 & 0.578 & 0.416 & 0.566 & 0.705 & 0.748 & 0.355 & \textcolor[rgb]{1.000, 0.000, 0.000}{\textbf{0.851}} \\
		Let   & 0.589 & 0.371 & 0.544 & 0.459 & 0.286 & 0.593 & 0.561 & 0.602 & 0.222 & \textcolor[rgb]{1.000, 0.000, 0.000}{\textbf{0.624}} \\
		Cad   & 0.370  & 0.687 & 0.660  & 0.628 & 0.683 & 0.377 & 0.554 & 0.465 & 0.763 & \textcolor[rgb]{1.000, 0.000, 0.000}{\textbf{0.959}} \\
		Mus   & 0.394 & \textcolor[rgb]{1.000, 0.000, 0.000}{\textbf{1.000}} & 0.527 & \textcolor[rgb]{1.000, 0.000, 0.000}{\textbf{1.000}} & 0.968 & 0.416 & 0.508 & 0.182 & 0.034 & 0.998 \\
		Opt   & 0.113 & 0.812 & 0.721 & 0.000     & 0.133 & 0.123 & 0.000     & 0.033 & 0.130  & \textcolor[rgb]{1.000, 0.000, 0.000}{\textbf{0.992}} \\
		Sat2  & 0.205 & 0.967 & 0.416 & \textcolor[rgb]{1.000, 0.000, 0.000}{\textbf{0.97}} & 0.936 & 0.209 & 0.614 & 0.453 & 0.821 & 0.954 \\
		Sat   & 0.529 & 0.648 & 0.538 & 0.685 & 0.674 & 0.529 & 0.620  & 0.529 & \textcolor[rgb]{1.000, 0.000, 0.000}{\textbf{0.816}} & 0.814 \\
		Pen   & 0.246 & 0.567 & 0.666 & 0.468 & 0.580  & 0.233 & 0.308 & 0.277 & 0.238 & \textcolor[rgb]{1.000, 0.000, 0.000}{\textbf{0.981}} \\
		Ann   & 0.394 & 0.290  & 0.383 & 0.324 & 0.393 & 0.347 & 0.368 & 0.361 & 0.378 & \textcolor[rgb]{1.000, 0.000, 0.000}{\textbf{0.896}} \\
		Mni   & 0.558 & 0.609 & 0.584 & 0.617 & 0.520  & 0.555 & 0.629 & 0.576 & 0.437 & \textcolor[rgb]{1.000, 0.000, 0.000}{\textbf{0.961}} \\
		Cam   & 0.463 & 0.581 & 0.537 & 0.580  & 0.547 & 0.383 & 0.541 & 0.537 & 0.610  & \textcolor[rgb]{1.000, 0.000, 0.000}{\textbf{0.851}} \\
		Shu   & 0.364 & 0.975 & 0.658 & 0.516 & 0.976 & 0.256 & 0.453 & 0.431 & 0.973 & \textcolor[rgb]{1.000, 0.000, 0.000}{\textbf{0.981}} \\
		Cel   & 0.107 & 0.386 & \textcolor[rgb]{1.000, 0.000, 0.000}{\textbf{0.640}} & 0.360  & 0.315 & 0.142 & 0.237 & 0.412 & 0.424 & 0.630 \\
		Fra   & 0.000     & 0.275 & 0.619 & 0.482 & 0.519 & 0.163 & 0.326 & 0.437 & 0.477 & \textcolor[rgb]{1.000, 0.000, 0.000}{\textbf{0.919}} \\
		Don   & 0.412 & 0.401 & 0.551 & 0.355 & 0.294 & 0.439 & 0.449 & 0.000     & 0.432 & \textcolor[rgb]{1.000, 0.000, 0.000}{\textbf{0.999}} \\
		Avg   & 0.417 & 0.609 & 0.571 & 0.551 & 0.563 & 0.419 & 0.500   & 0.450  & 0.485 & \textcolor[rgb]{1.000, 0.000, 0.000}{\textbf{0.840}} \\
		Rank  & 7.2   & 4.95  & 5.28  & 5.35  & 5.1   & 7.3   & 5.48  & 6.48  & 6.13  & 1.75 \\
		\bottomrule[1.5pt]
	\end{tabular}%
	\label{table_4}%
\end{table*}%

\begin{table}[htbp]
	\centering
	\caption{The relative performance gains of EAL-GAN over its three best competitors on small and large datasets.}
	\begin{tabular}{cccccc}
		\toprule[1.5pt]
		& \multicolumn{2}{c}{AUC gains (\%)} &       & \multicolumn{2}{c}{${Gmean}$ gains (\%)} \\
		\cmidrule{2-3}\cmidrule{5-6}         & Small & Large &       & Small & Large \\
		\cmidrule{1-3}\cmidrule{5-6}    \emph{vs.}CBLOF & +10.24 & +20.66 &       & +33.38 & +66.24 \\
		\emph{vs.}DevNet & +4.35  & +7.32  &       & +34.29 & +58.65 \\
		\emph{vs.}iForest & +15.18 & +18.47 &       & +41.66 & +52.07 \\
		\bottomrule[1.5pt]
	\end{tabular}%
	\label{table_5}%
\end{table}%

\begin{table}[htbp]
	\centering
	\caption{Wilcoxon testing result between EAL-GAN and DevNet (AUC as the measure)}
	\begin{tabular}{rrrrr}
		\toprule[1.5pt]
		\multicolumn{1}{l}{Comparison} & \multicolumn{1}{l}{\emph{R}+} & \multicolumn{1}{l}{\emph{R}-} & \multicolumn{1}{c}{\emph{p}-value} &
		\multicolumn{1}{l}{Hypothesis(0.05)} \\
		\midrule[1.5pt]
		CB-GAN \emph{vs.} DevNet & 190   & 0     & ${0.000 < 0.05}$ & \multicolumn{1}{c} {\textbf{Rejected}} \\
		\bottomrule[1.5pt]
	\end{tabular}%
	\label{table_6}%
\end{table}

\textbf{Overall Evaluation}. According to the results in Table 3 and 4, it is evident that the proposed EAL-GAN outperforms its competitors in terms of both AUC and ${Gmean}$. In particular, when taking AUC as the performance measure, EAL-GAN achieves the highest AUC on 14 out of the 20 datasets. It is clear that the proposed EAL-GAN achieves substantially better average AUC than the best performed traditional anomaly detector CBLOF (16.4\%), the best ensemble detector iForest (16.8\%) and the state-of-the-art deep anomaly detector DevNet (5.67\%). Other classical detection methods have poorer performances. In particular, LOF and FB perform the least well. Actually, a relatively good performance has been observed when adopting LOF and FB on small datasets. However, a significant performance drop has been observed when adopting them on large-scale datasets, suggesting the traditional anomaly detectors do not scale to large datasets. The inferior performance of FB might be caused by the fact that when sufficient data are given to train the model, it is difficult for unsupervised methods to construct diverse members. In contrast, a novel loss function is proposed in EAL-GAN to promote complementariness within its members. Other methods (e.g., AvgKNN, FOD, MOGAAL) provide better performance than LOF and FB, but compared with the proposed EAL-GAN, their performances are not as good.

When taking ${Gmean}$ as the performance measure, the superiority of EAL-GAN over other methods is even more pronounced. EAL-GAN achieves markedly better average ${Gmean}$ across the 20 datasets than the best traditional anomaly detector CBLOF (52.5\%), the best ensemble detector iForest (49.2\%), and the best deep anomaly detector DSVDD (37.9\%). All these results reveal a strong capability of EAL-GAN in leveraging limited prior information to construct effective anomaly detectors in various tasks.

\textbf{Further Analysis}. The performance of a GAN heavily depends on the size of training data. To explore the influence of data size on the proposed EAL-GAN, we examine the model performance on datasets exhibiting different data size. In particular, we further analyze EAL-GAN and three of its best competitors (i.e., CBLOF, DevNet and iForest). To that end, we split the 20 datasets into two groups at the data size of 3000: small-data (the first 8 datasets in Table 1 whose data sizes are less than 3000) and large-data (the remaining 12 datasets in Table 1). The relative performance gains are reported in Table 5. Table 5 shows that: (i) EAL-GAN improves the performance over its competitors on both small and large datasets, and the relative performance gains on large datasets are higher than those on small datasets. This suggests that EAL-GAN is better for large-scale datasets, where more data can be used to boost the model. (ii) Among the three best competitors, DevNet is the best performer. It is worth mentioning that both DevNet and EAL-GAN utilize a small set of labeled data as prior knowledge. This suggests this extra knowledge can improve the model and largely help them to be resilient to the noises in the training data. By contrast, the unsupervised methods, such as CBLOF and iForest, don’t have any labeled information and their performance decreases with increasing in data size, because the unsupervised methods generally assume that anomalies in the dataset are rare and diverse, thus they perform inferiorly when the increasing in absolute number of anomalies violates the assumption. 

\begin{figure}[]
	\centering
	\subfloat[AUC as the measure]{\includegraphics[width=7.5cm, height=3cm]{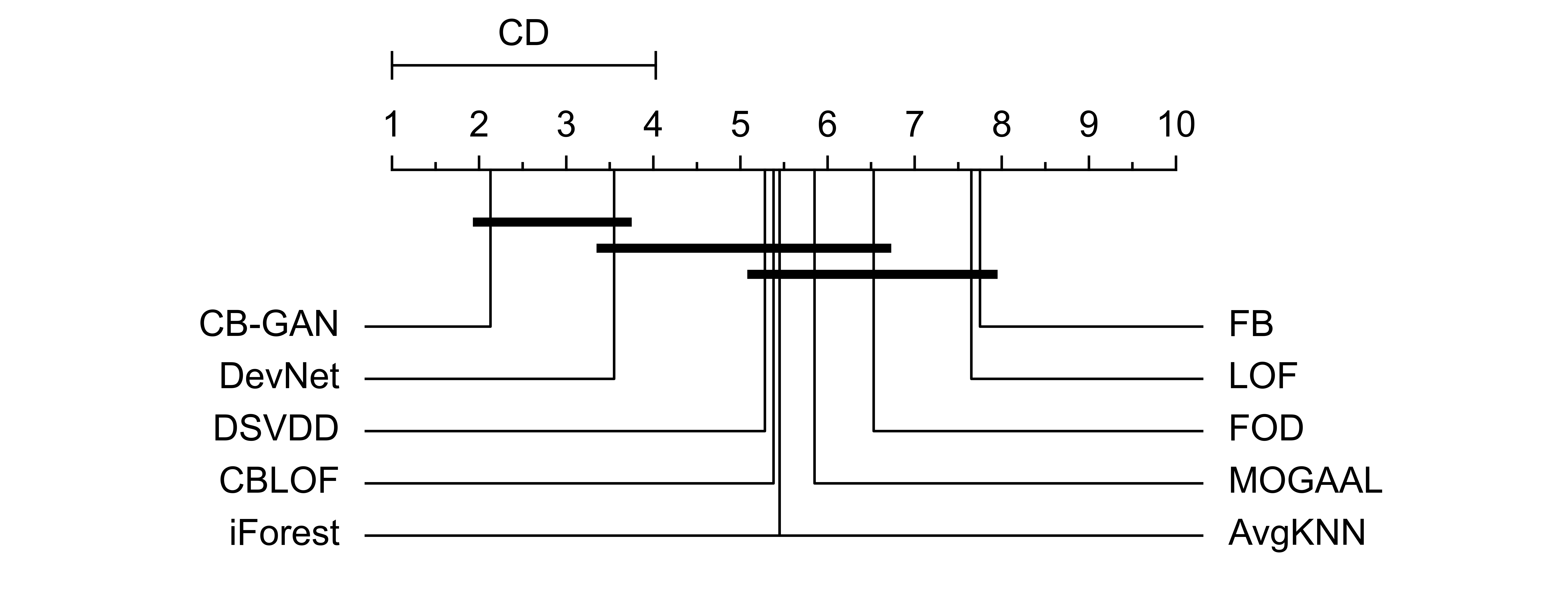}
		\label{fig_4a_case}}
	\hfil
	\subfloat[Gmean as the measure]{\includegraphics[width=7.5cm, height=3cm]{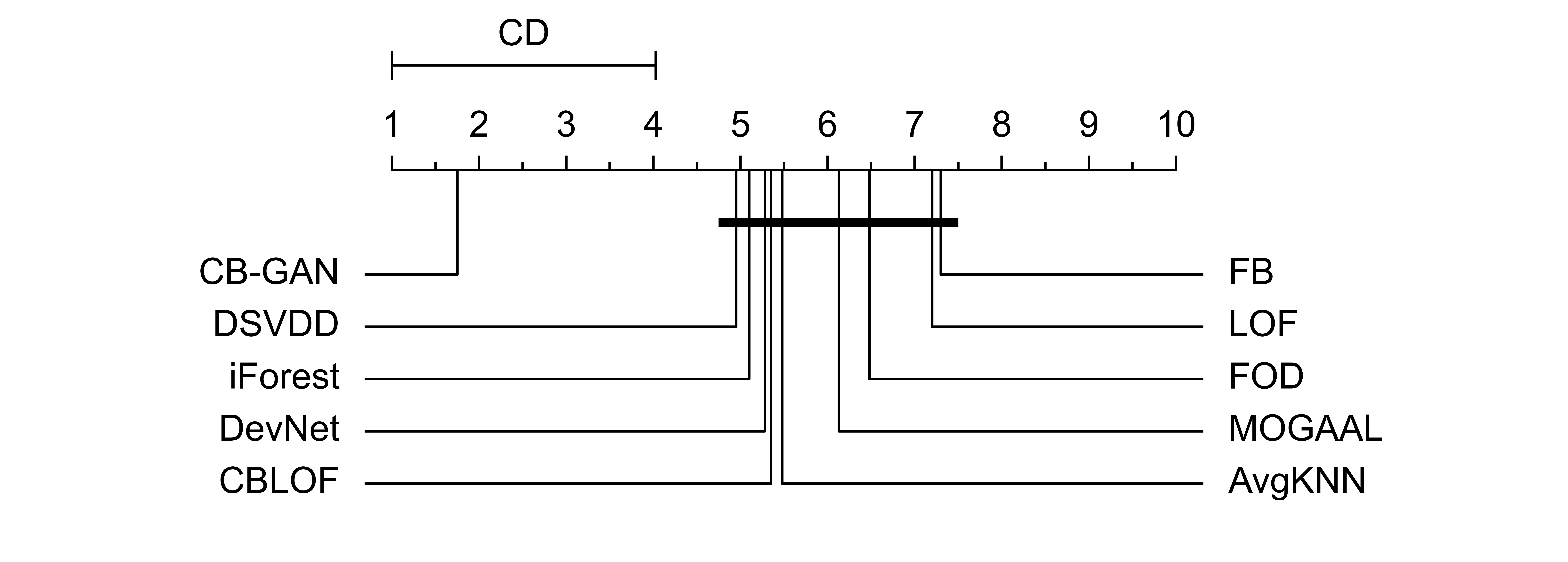}
		\label{fig_4b_case}}
	\hfil
	
	\caption{Critical difference diagrams for the Nemenyi test. The average ranking for each method on 20 datasets are shown, with ranking 1 indicating the most accurate method. Horizontal line segments group together methods with rankings that are not significantly different.}
	\label{fig_4}
\end{figure}

\textbf{Statistical test}. The advantages of EAL-GAN can also be confirmed by the statistical test results. Friedman test assumes that there exists no significant difference between all the comparison methods. In other words, the average rankings ${{R_j} = \sum\nolimits_{i = 1}^N {r_i^j}}$ of \emph{k} methods on \emph{N} datasets should be similar to each other, where ${r_i^j}$ denotes the ranking of \emph{j}-th method on the \emph{i}-th dataset, and ${r_i^j=1}$ if \emph{j}-th method provides the best performance on the \emph{i}-th dataset. Thus, a lower average ranking ${R_j}$ indicates a better anomaly detection ability. Taking the average rankings in Table 3 and 4 as input, the Friedman test outputs ${p=0.00<0.05}$ and ${p = 0.00 < 0.05}$ as the statistical test results, which are low enough to reject the null hypothesis of Friedman test. Thereafter, we continue with Nemenyi test. The critical difference (\emph{CD}) of Nemenyi test for comparing 10 methods on 20 datasets are calculated as follows:

\begin{equation}
{CD = {q_a}\sqrt {\frac{{k(k + 1)}}{{6N}}}  = {q_a}\sqrt {\frac{{10 \times (10 + 1)}}{{6 \times 20}}} }
\end{equation}

The critical value of ${q_a}$ for ${\alpha  = 0.05}$ is 3.164 and the corresponding \emph{CD} is 3.03. The critical difference diagrams for the Nemenyi test are plotted in Fig.4, where AUC and ${Gmean}$ are taken as the measures. From Fig. 4, several observations can be made: (i) No matter AUC or ${Gmean}$ is taken as the performance measure, EAL-GAN is significantly better than all the other methods at the 95\% confidence level, except in one case, i.e., EAL-GAN \emph{vs.} DevNet in Fig.4a. As pointed out by \cite{RN63}, sometimes Friedmen test detects a significant difference, but its post-hoc test may fail to detect it, due to the lower capability of the latter. Thus, we supplement the Nemenyi test with Wilcoxon signed-ranking test on the comparison between EAL-GAN and DevNet. The Wilcoxon test results are reported in Table 6, which confirms that the difference between EAL-GAN and DevNet is statistically significant. (ii) The statistical test results also confirm that DevNet is statistically better than LOF and FB. There exists no significant difference between the remaining methods. 

With all the results reported above, we can state that improvements of EAL-GAN over its competitive methods are statistically significant.

\subsection{Experiments on Parameter Setting}
EAL-GAN involves multiple hyper-parameters. In this part, we study the hyper-parameters which we believe are vital for the EAL-GAN (i.e., the number of discriminators, the depths of generator, the ratio of sampling) and investigate the robustness of EAL-GAN against various parameter settings.

The number of discriminators plays a vital role in the proposed EAL-GAN. A sufficient number of discriminators are expected to inject more intense competition into the GAN training procedure, therefore lead to a better anomaly detection capability. To verify whether this goal is achieved, we varied the number of discriminators in the range of 1 to 21, and then illustrated the performance of EAL-GAN with different number of discriminators in Fig.5a and Fig.5b. It can be observed that the average performance of the model across 20 datasets is improved after multiple discriminators are adopted. This observation is rather consistent no matter AUC or ${Gmean}$ is taken as the measure, suggesting that multiple discriminators can bring better generalization. 

The depth of the generator is another important hyper-parameter for EAL-GAN. A higher depth indicates higher capacity of neural network. To investigate the influence of network depth, we adjusted the number of layers in the generator from 3 to 5, and results are shown in Fig.5c and Fig.5d. It can be seen that 4-layer generator achieves slightly better detection ability than 3-layer generator (i.e., relative improvements are 0.9\% when AUC is the measure) and 5-layer generator (i.e., relative improvements are 0.5\%). The relative worse performance of 3-layer generator is mainly caused by the lower capacity of the neural network structure. The 5-layer generator encounters the vanishing gradient problem, a common problem in deep neural network training. Actually, in our preliminary experiments, we have tried to build generators with deeper structure (e.g., 6-layer and 7-layer), but these models keep failing on datasets with high dimension, such as \emph{Arrhythmia} and \emph{Musk}, whose data dimensions are 274 and 166. 

As for the sampling ratio, a higher ratio can bring more priori knowledge to the training process. We adjusted the sampling ratio in the range of 5\% to 95\% with a step size of 10\%, and the results are given in Fig.5e and Fig.5f. It can be seen that as the sampling ratio increases, the average AUC of model increases from 0.951 to 0.977 and remains relatively stable once the ratio reaches a certain level. As the sampling ratio increases, more normal data will be introduced into the training set and the training set can exhibit more severely imbalanced distribution. EAL-GAN constantly provides excellent performance in such detrimental scenarios, suggesting EAL-GAN is robust against various level of contamination.

\begin{figure}[]
	\centering
	\subfloat[]{\includegraphics[width=1.7in, height=1.3in]{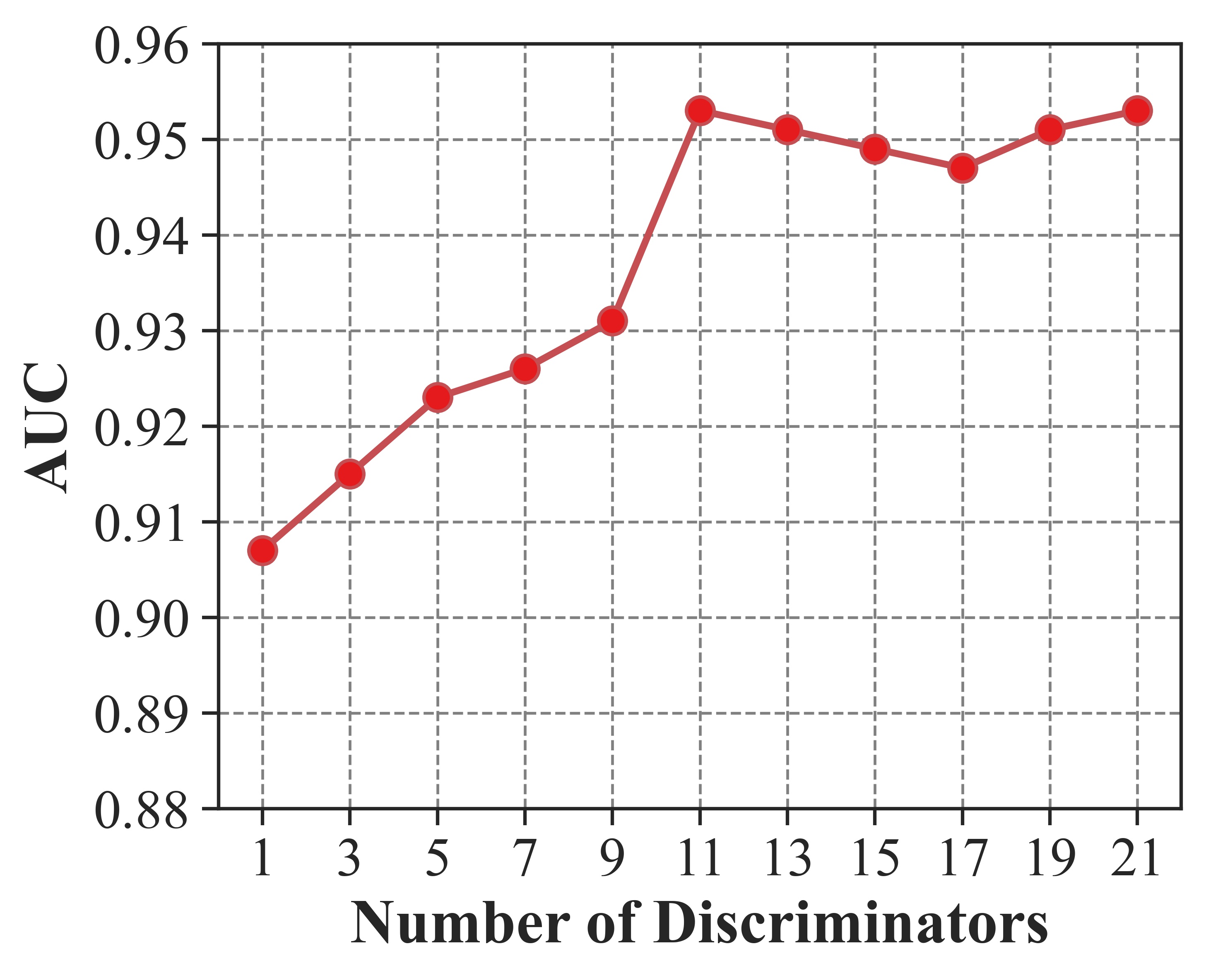}
		\label{fig_5a}}
	\hfil
	\subfloat[]{\includegraphics[width=1.7in, height=1.3in]{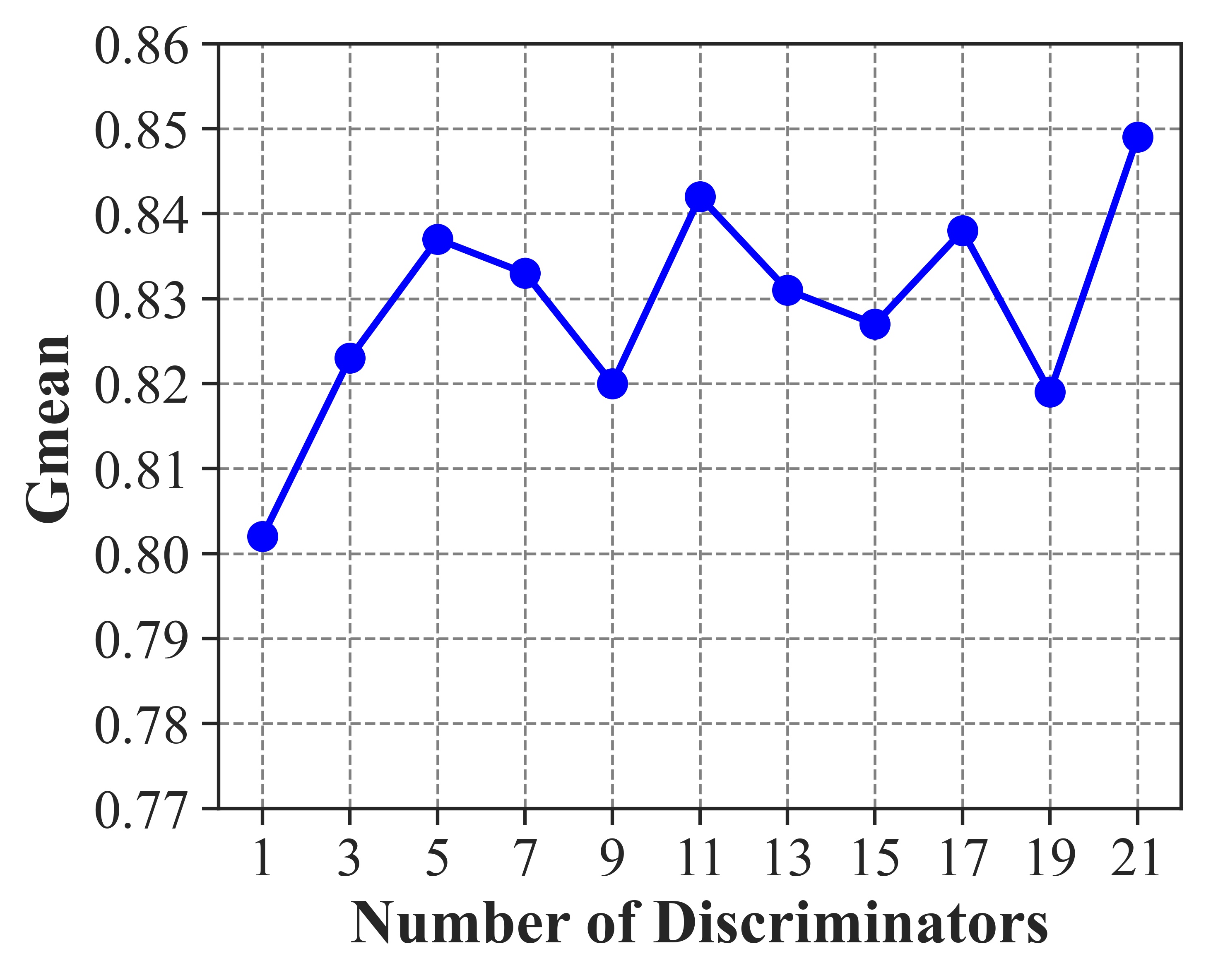}
		\label{fig_5b}}
	\hfil
	
	\subfloat[]{\includegraphics[width=1.7in, height=1.3in]{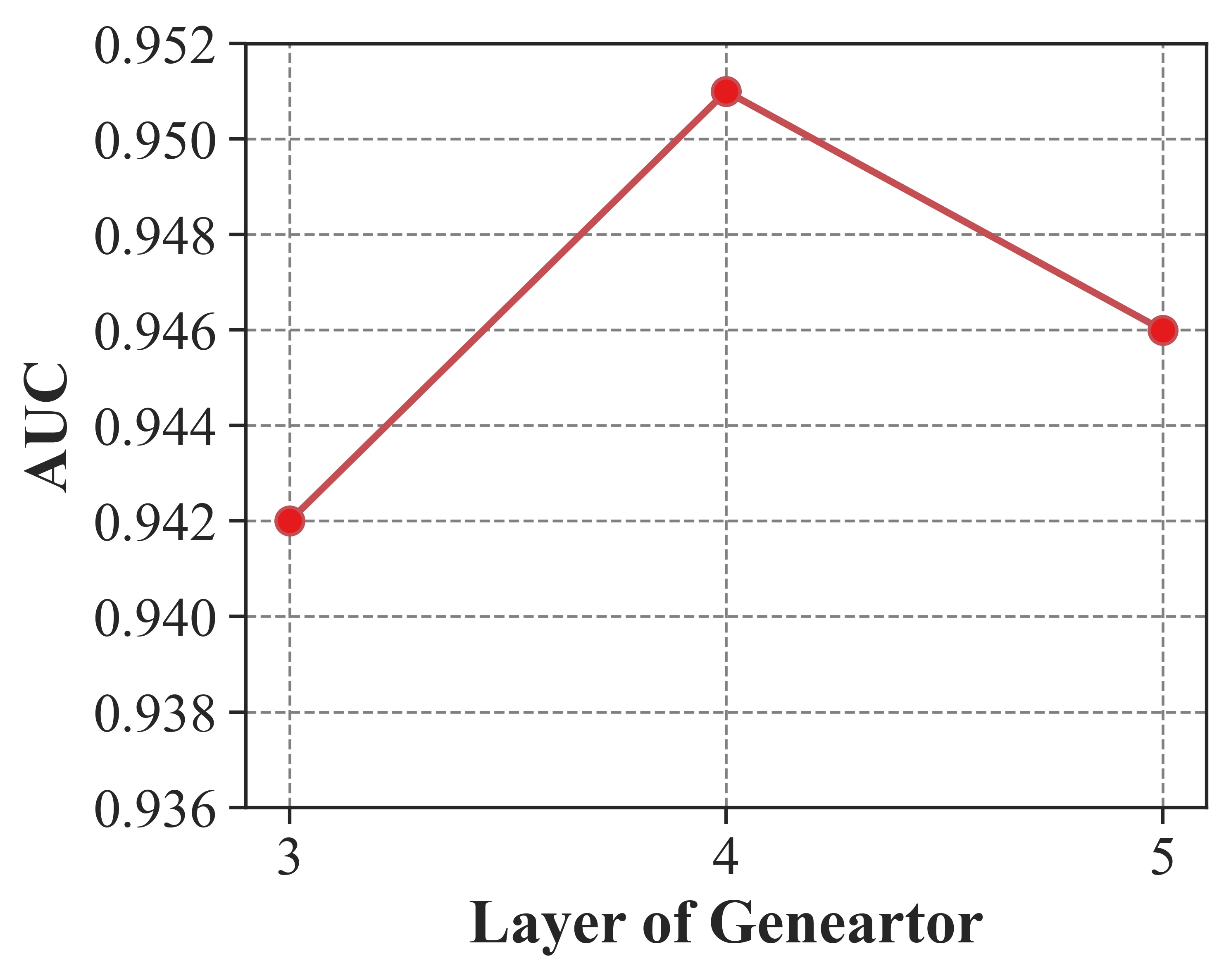}
		\label{fig_5c}}
	\hfil
	\subfloat[]{\includegraphics[width=1.7in, height=1.3in]{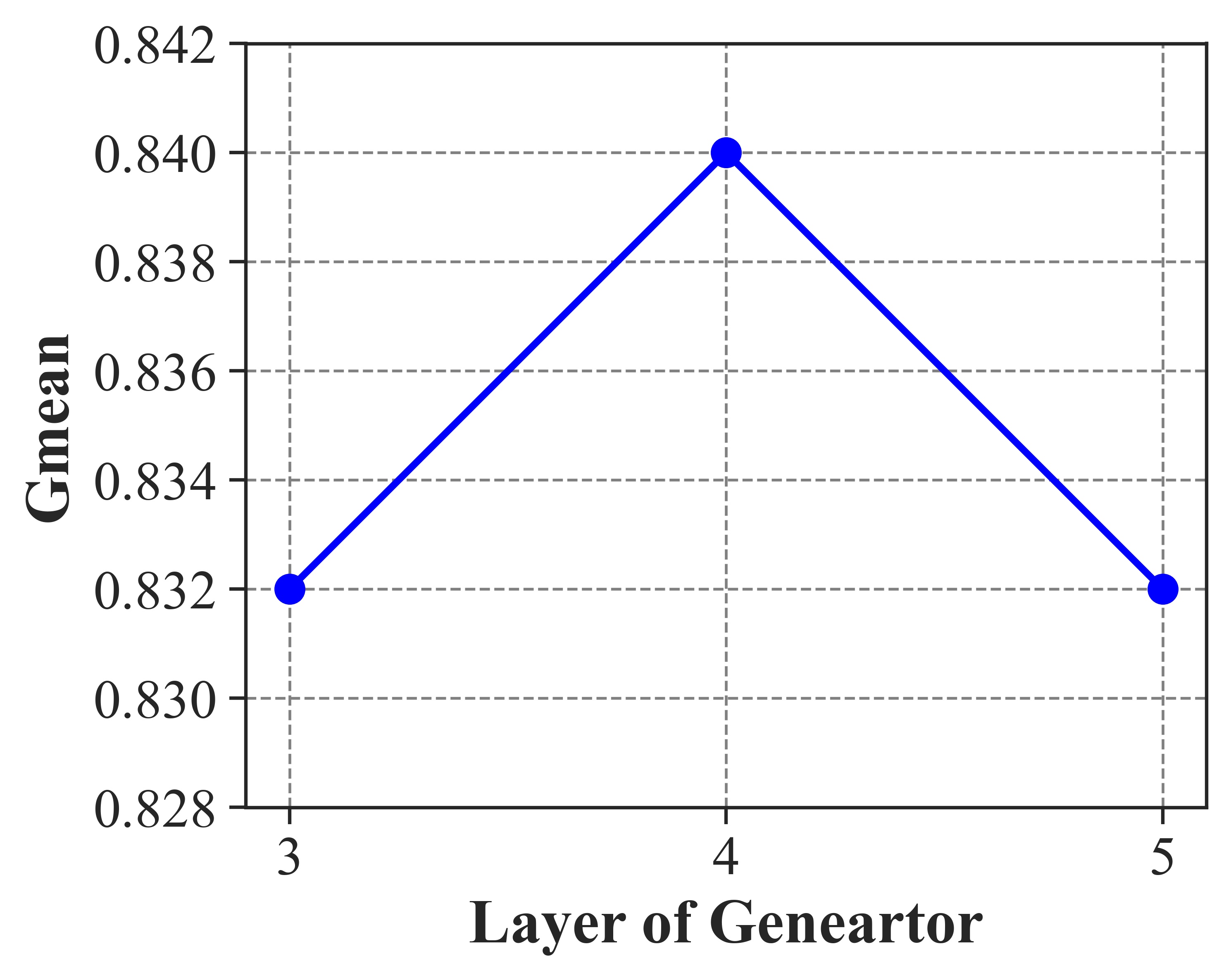}
		\label{fig_5d}}
	\hfil
	\subfloat[]{\includegraphics[width=1.7in, height=1.3in]{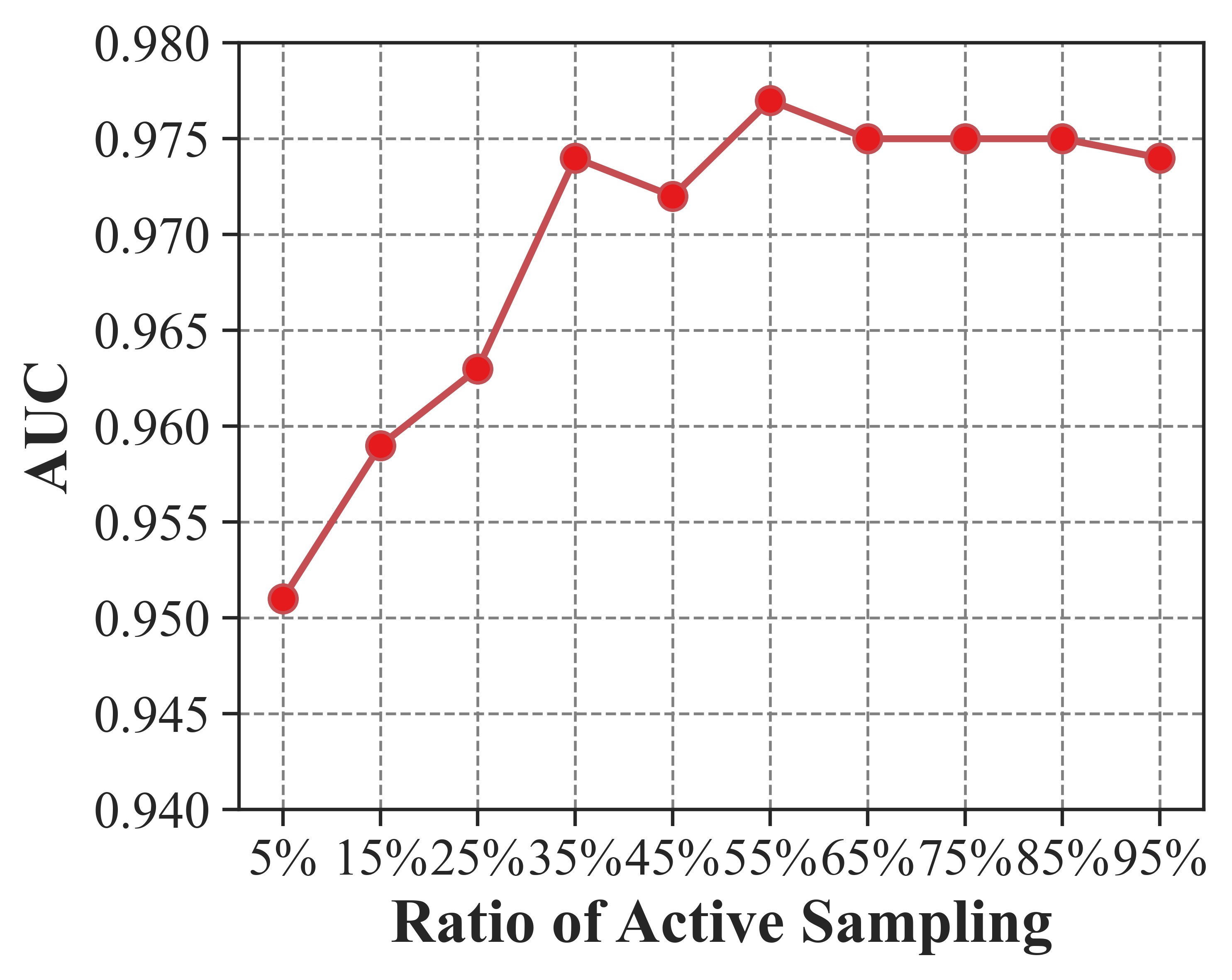}
		\label{fig_5e}}
	\hfil
	\subfloat[]{\includegraphics[width=1.7in, height=1.3in]{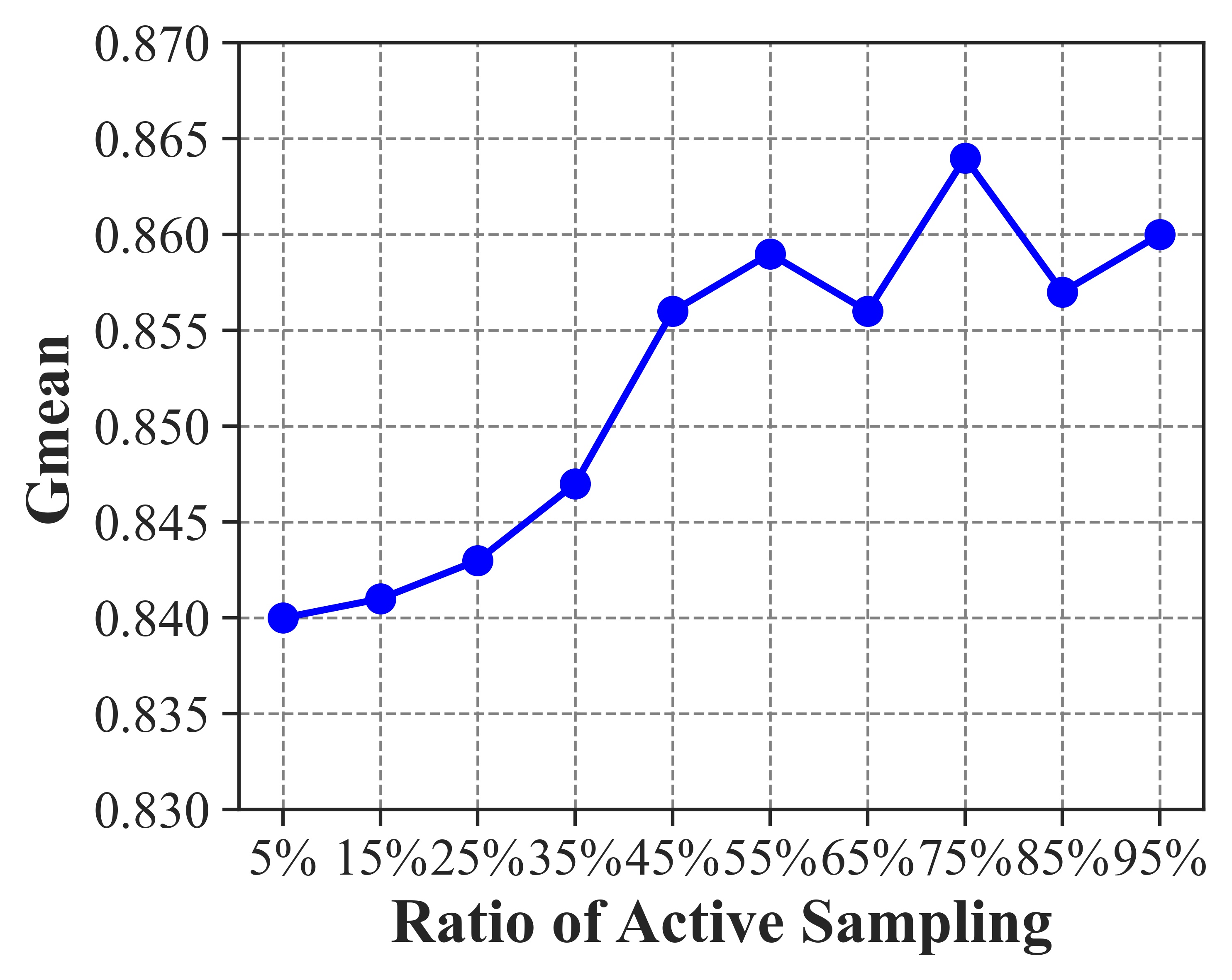}
		\label{fig_5f}}
	\hfil
	\caption{The influence of different parameter settings on the proposed EAL-GAN. The red lines indicate that AUC is taken as the performance measure, while blue lines are for the ${Gmean}$.}
	\label{fig_5}
\end{figure}

\subsection{The Quality and Usefulness of the Generated Data}
Theoretically, the generator can provide unlimited number of fake data to train the anomaly detector. Thus, a natural question arises: Can these unlimited number of fake data unlimitedly boost the anomaly detector? To answer this question, we train the anomaly detector with different number of generated data, and observe how its performance changes.

\begin{table*}[!htbp]
	\centering
	\caption{The performance of EAL-GAN under different Fake:Real ratio. Fake:Real=1:0 indicates that only fake data is used to train the anomaly detector.}
	\begin{tabular}{rrrrrrrrr}
		\toprule[1.5pt]
		Ratio & \multicolumn{1}{c}{1:0} & \multicolumn{1}{c}{0:1} & \multicolumn{1}{c}{1:1} & \multicolumn{1}{c}{5:1} & \multicolumn{1}{c}{10:1} & \multicolumn{1}{c}{20:1} & \multicolumn{1}{c}{50:1} & \multicolumn{1}{c}{100:1} \\
		\midrule[1.5pt]
		AUC   & 0.906 & 0.911 & 0.938 & 0.943 & 0.946 & 0.952 & \textcolor[rgb]{1.000, 0.000, 0.000}{\textbf{0.958}} & 0.955 \\
		Gmean & 0.804 & 0.808 & 0.833 & 0.838 & 0.825 & 0.84  & \textcolor[rgb]{1.000, 0.000, 0.000}{\textbf{0.842}} & 0.841 \\
		\bottomrule[1.5pt]
	\end{tabular}%
	\label{table_7}
\end{table*}%

To this end, when optimizing the auxiliary classifiers, we change the amount of generated data, so that the ratio between the number of fake samples and that of real data varies. All the other parameters are kept the same as the ones in Section 4.2, i.e., the number of fake data and real data used to update the generator and adversarial classifiers haven’t been changed. The average performance of EAL-GAN on 20 real-world datasets are reported in Table 7 and it is shown that:

(1) When only fake data is used to train the anomaly detectors (i.e., Fake:Real=1:0), it achieves an average AUC of 0.906, which is only 0.5\% lower than 0.911 achieved by the detectors trained with purely real data (i.e., Fake:Real=0:1), it achieves an average ${Gmean}$ of 0.804, which is also about 0.5\% lower than 0.808 achieved by the detectors trained with purely real data. This shows that EAL-GAN is a very high-quality conditional data generator in the sense that its generated samples are as good as real data in training anomaly detectors. In other words, EAL-GAN capture the distributions of the real data extremely well. Interestingly, our detectors trained based either on purely generated data or purely real data have better averaging AUC than the best competitor DevNet (0.906 and 0.911 \emph{vs.} 0.900) and significantly better averaging ${Gmean}$ than the best competitor DSVDD (0.804 and 0.808 \emph{vs.} 0.609).

(2) The models trained with both fake and real data provide significantly better performances than the model trained with only fake data (e.g., when changing the Fake:Real ratio from 1:0 to 1:1, the relative improvements of average AUC and ${Gmean}$ are 3.5\% and 3.6\%, respectively). It's worth noting that the fake data are generated from random input noises without any extra intervention, therefore, the fake data can exhibit diverse quality. On the other hand, the real data are selected by the proposed ensemble active learning strategy. The performance gain brought by introducing the real data not only demonstrates the effectiveness of the proposed active learning in identifying the most informative data, it also suggests the technique like Top-K training \cite{RN64}, which throws away the bad generated data during GAN training, can further improve the proposed method.

(3) The fake data can’t unlimitedly boost the anomaly detector. In particular, when the Fake:Real ratio is under 100:1, training the anomaly detector with more fake data can almost always improve the performance of the final detector. When the Fake:Real ratio is above 100:1, a slight performance drop is observed. A possible reason is that EAL-GAN is supported by limited number of real data, therefore, the diversity of generated data is limited. However, this observation well explains our major motivation of proposing EAL-GAN: By devoting a small amount of effort to annotate the real data, EAL-GAN can generate significantly more labeled data than the available real data, and these generated data can help boost the anomaly detector, providing a promising solution to the lack of labeled data problem for a wide range of real-world applications.

\subsection{Ablation Study}
In this subsection, we examine the importance of the key components of EAL-GAN by comparing EAL-GAN with its four variants. 

\begin{enumerate}
	\itemsep=0pt
	\item \textbf{EAL-GAN-emb}, which removes the embedding layer in the discriminator network (please see Figure 1). In this case, the discriminator architecture of EAL-GAN-emb would be identical to the one in AC-GAN. 
	\item \textbf{EAL-GAN-single}, which adopts only one discriminator without any ensemble. 
	\item \textbf{EAL-GAN-loss}, which replaces the loss functions in equation (4) and (7) with standard adversarial loss. 
	\item \textbf{EAL-GAN-random}, which replaces the active sampling procedure in EAL-GAN with a random sampling procedure. EAL-GAN-random randomly selects the same number of unlabeled data as the input for EAL-GAN and then manually labels them.
	
\end{enumerate}

As for the datasets, synthetic datasets described in Table 2 are generated. We run EAL-GAN and its four variants 10 times on each of the synthetic datasets. In each run, 60\% of the available data are randomly selected as the training set and the remaining 40\% data are taken as the testing set. The AUC and ${Gmean}$ obtained from all the testing sets are averaged into one single measure, and the results are shown in Fig.6. It can be observed that almost all the methods can provide good performance on the single-cluster data, except the EAL-GAN-single method. As the data distributions get more complex, all the four variants provide significantly worse performances, suggesting all the components make their contributions towards improving the EAL-GAN.

\begin{figure}
	\centering
	\subfloat[AUC is taken as the performance measure]{\includegraphics[width=3in, height=1.6in]{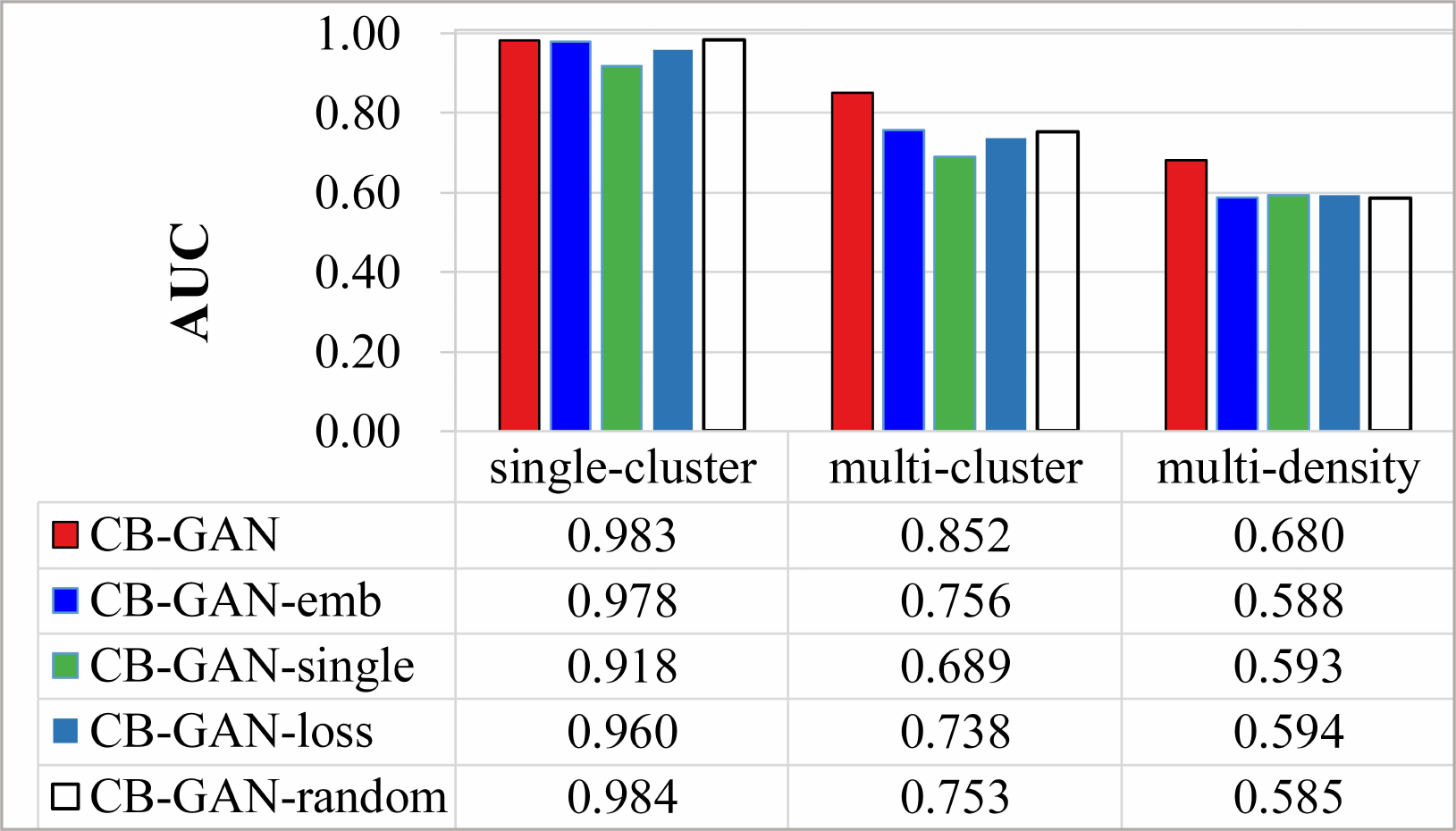}
		\label{fig_6a}}
	\hfil
	\subfloat[${Gmean}$ is taken as the performance measure]{\includegraphics[width=3in, height=1.6in]{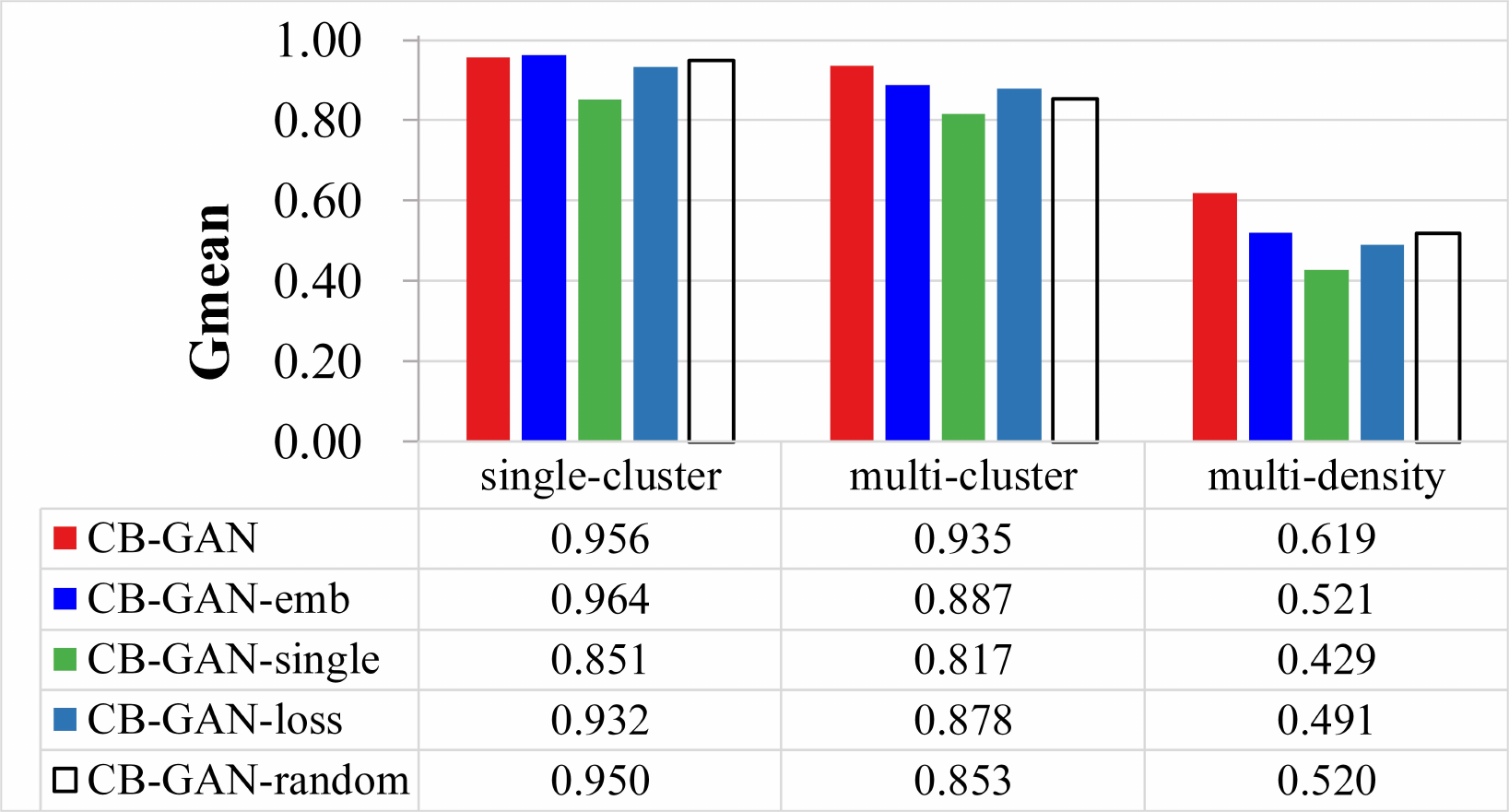}
		\label{fig_6b}}
   \caption{The performance of EAL-GAN and its four variants on synthetic datasets.}
   \label{fig_6}
\end{figure}

\begin{figure}[!htbp]
	\centering
	\subfloat[EAL-GAN]{\label{Fig_7a}\includegraphics[width=1\linewidth]{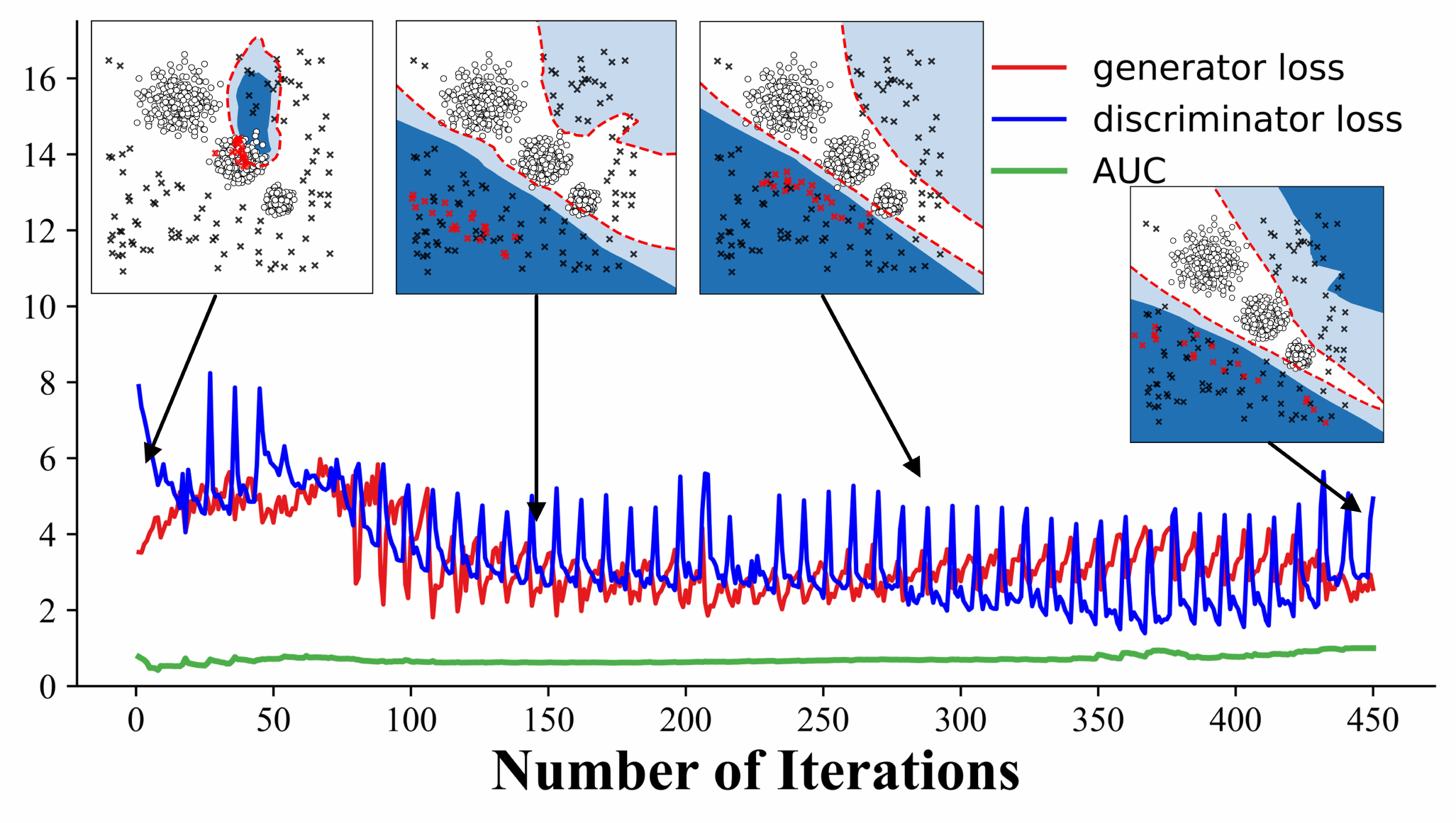}}
	\hfil
	
	\subfloat[EAL-GAN-emb]{\label{Fig_7b}\includegraphics[width=1\linewidth]{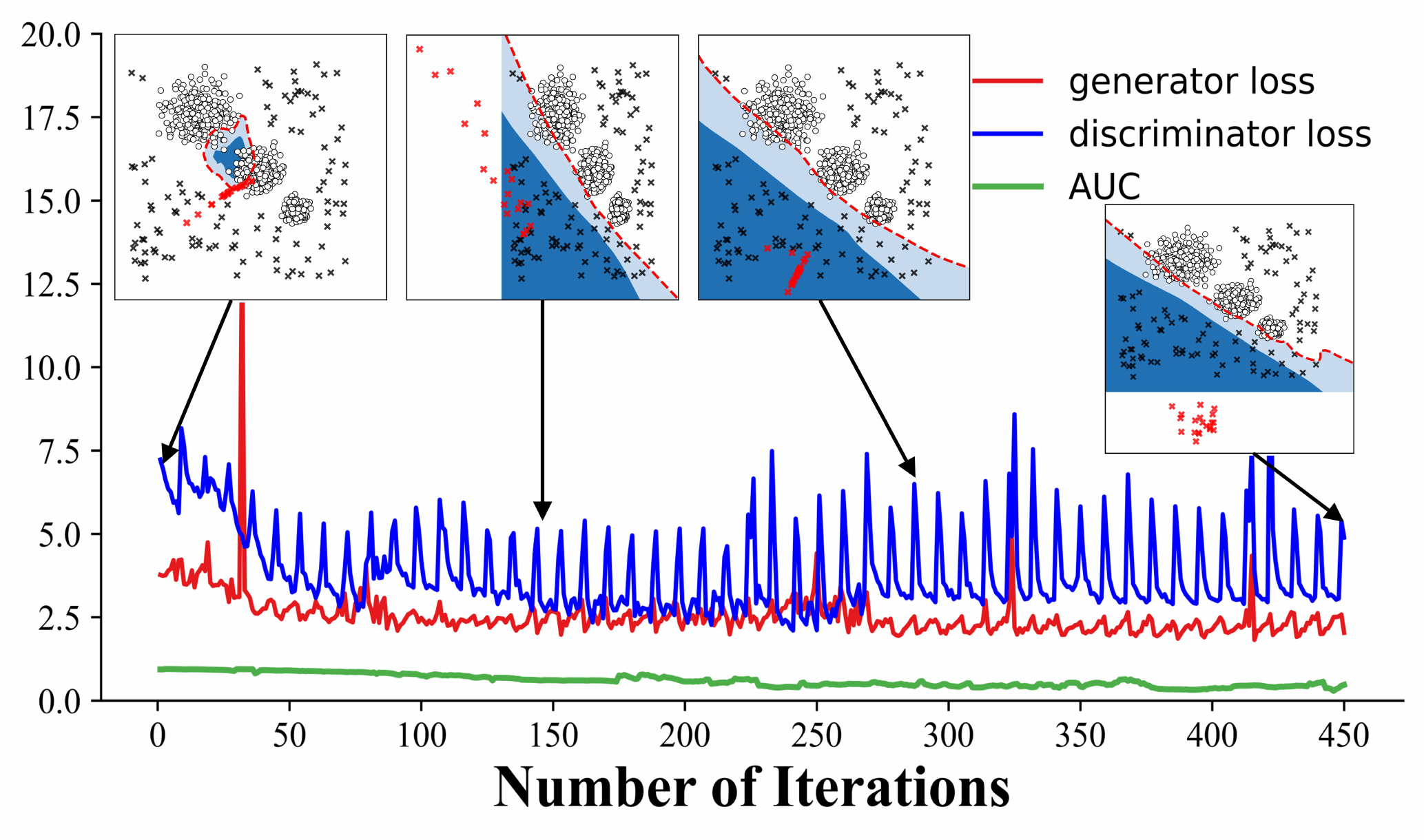}}
	\hfil
	
	\subfloat[EAL-GAN-single]{\label{Fig_7c}\includegraphics[width=1\linewidth]{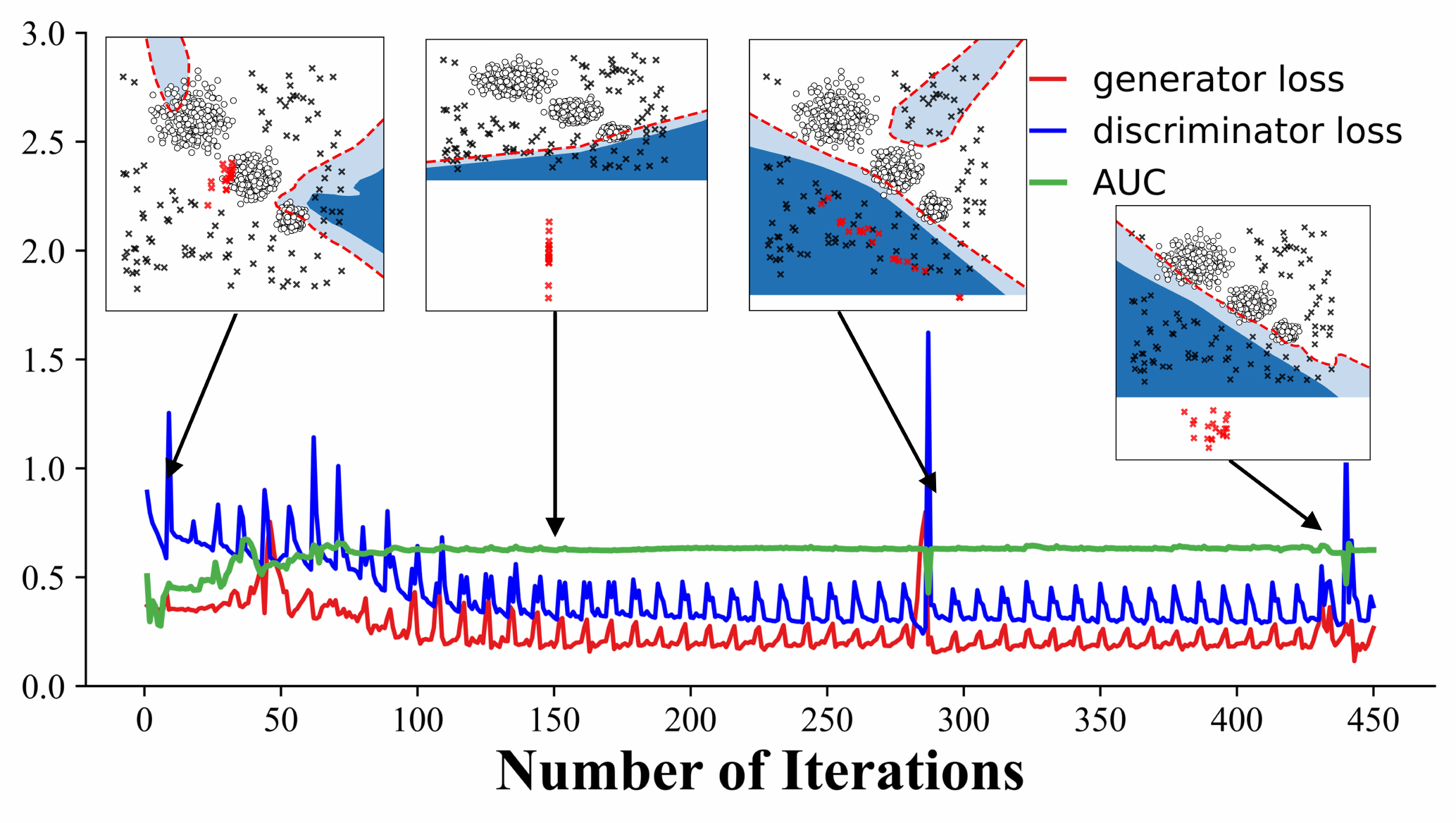}}
	\hfil
	\subfloat[EAL-GAN-loss]{\label{Fig_7d}\includegraphics[width=1\linewidth]{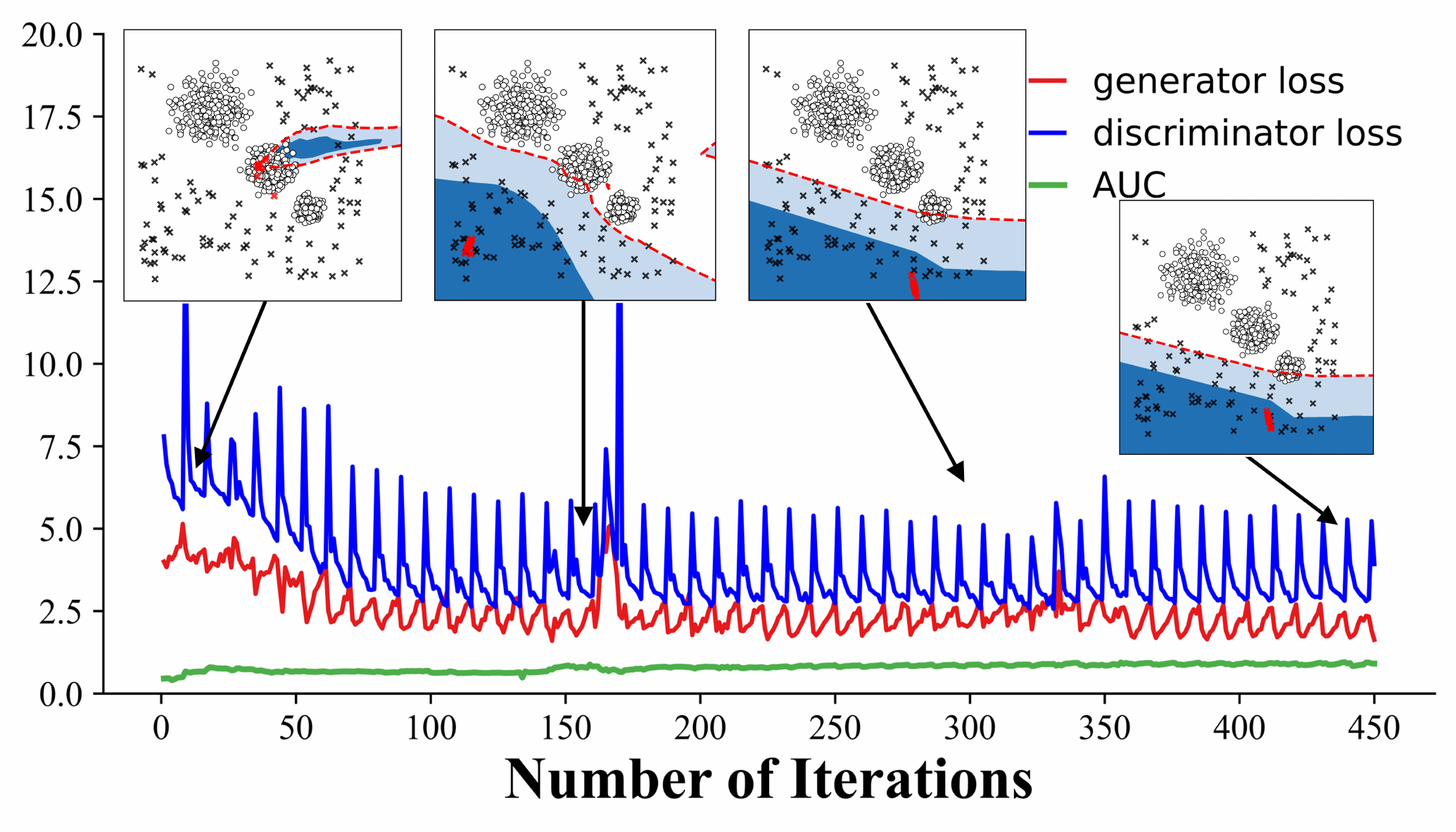}}
\end{figure}

\begin{figure}[!htbp]
	\addtocounter{subfigure}{4}
	%\ContinuedFloat  
	\subfloat[EAL-GAN-random]{\label{Fig_7e}\includegraphics[width=1\linewidth]{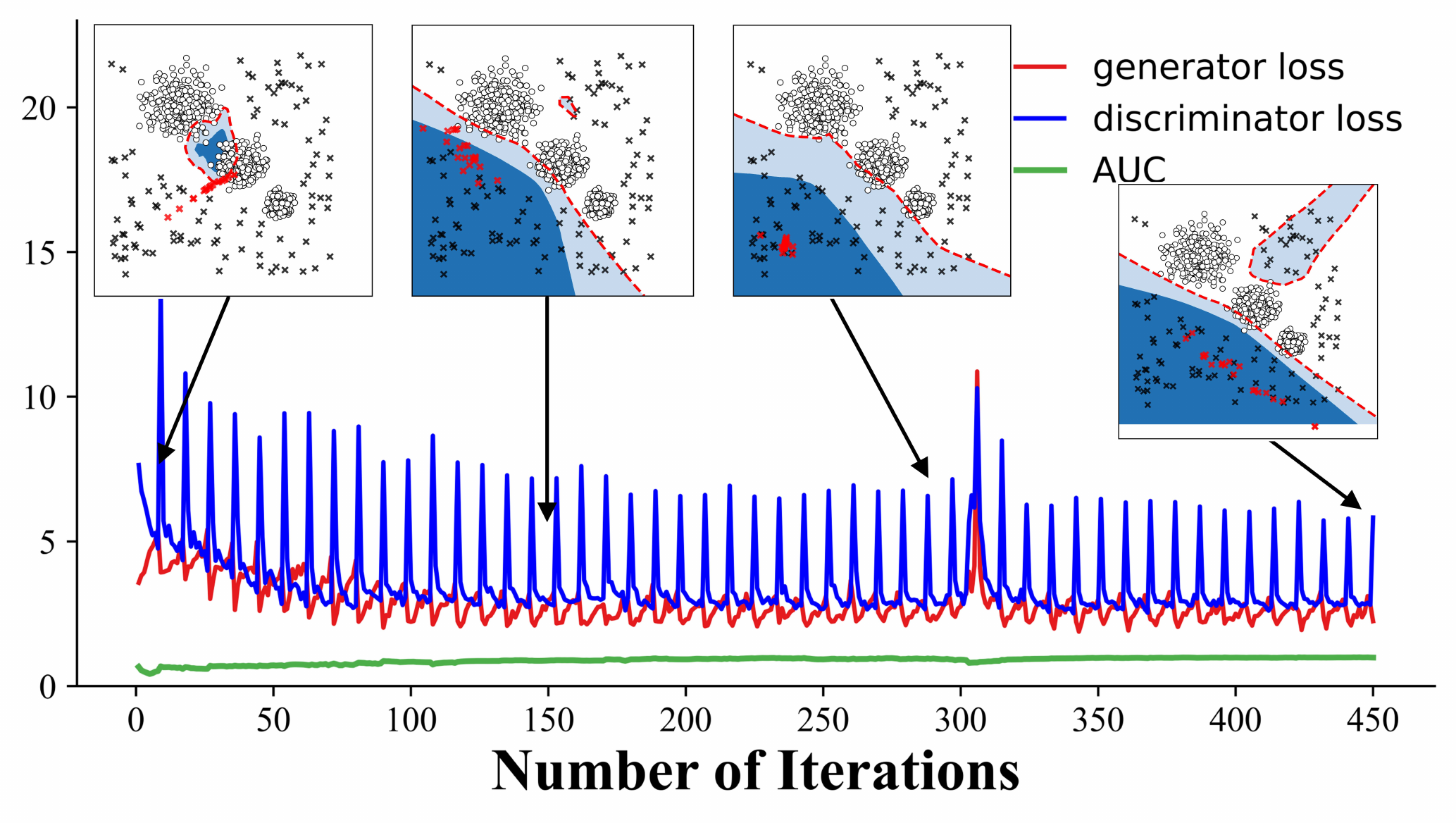}}
	
	\caption{The learning process of EAL-GAN and its four variants on a 2-D multi-density dataset (Note, these figures are best observed in an electronic copy by zooming in to clearly see the data points and decision boundaries). The generator loss is shown in red line, the discriminator loss with blue line and the AUC with the green line. The top four pictures in each sub-figure plot the classification boundary and the generated anomalies during the learning process.  In each picture, the hollow circle denotes the normal data, the black X symbol denotes the anomalies, the red X symbols are the generated anomalies. The red dash line is the division boundary provided by the detector. Data closer to the dark blue area are more likely to be anomies.}
	\label{fig_7}
\end{figure}

To further clarify the details, we provide the visual representations of the adversarial learning process of EAL-GAN and its variants on a synthetic 2-D multi-density dataset in Fig.7 (please zoom for better details). It can be clearly seen that as the adversarial learning proceed, EAL-GAN can progressively improve its classification boundary between normal data and anomalies. In addition, EAL-GAN can generate diverse anomalies near the decision boundary. On the other hand, EAL-GAN-emb and EAL-GAN-single can’t provide an accurate decision boundary and the generated anomalies are very different to the real anomalies. The anomalies generated by the EAL-GAN-loss are identical to each other. A plausible reason is that without the proposed loss function, all the discriminators exhibit similar behaviors, consequently, the generator can easily fool all the discriminators by generating similar data. Among the three variants, EAL-GAN-random is the best performer. EAL-GAN-random can generate diverse anomalies near the boundary. However, due to the random data selection procedure, EAL-GAN-random can’t obtain sufficient prior knowledge to learn the accurate classification boundary.

\section{Conclusions}
In this paper, we propose a supervised anomaly detector, called EAL-GAN, which is based on cGAN and designed to address the two major difficulties in developing anomaly detector. To tackle the problem of class imbalance, we introduce an ensemble of discriminators into cGAN and design a novel ensemble learning loss function that ensures the discriminators complement each other and the system is not dominated by the class with much more training data. To address the issue of lack of sufficient labeled training samples, we use cGAN to generate class balanced data to aid the design of the anomaly detector and introduce an active learning strategy to significantly reduce the burden of manually labeling training samples. 

Extensive experiments have been performed on real and synthetic datasets. The experimental results have comprehensively demonstrated the excellent anomaly detection performances of the proposed EAL-GAN and the necessity of incorporating all the components in EAL-GAN. For a long time, the anomaly detection task has been generally addressed by unsupervised methods, due to the lack of labeled information in the available datasets. Our scheme provides a data-efficient way to construct an accurate supervised detector from an imbalanced dataset.

\appendices

\ifCLASSOPTIONcompsoc
  \section*{Acknowledgments}
\else
  \section*{Acknowledgment}
\fi

This work has been submitted to the IEEE for possible publication. Copyright may be transferred without notice, after which this version may no longer be accessible

\ifCLASSOPTIONcaptionsoff
  \newpage
\fi



\bibliographystyle{IEEEtran}
\bibliography{EALGAN}
%\begin{thebibliography}{1}


%\end{thebibliography}

\begin{IEEEbiography}[{\includegraphics[width=1in,height=1.25in,clip,keepaspectratio]{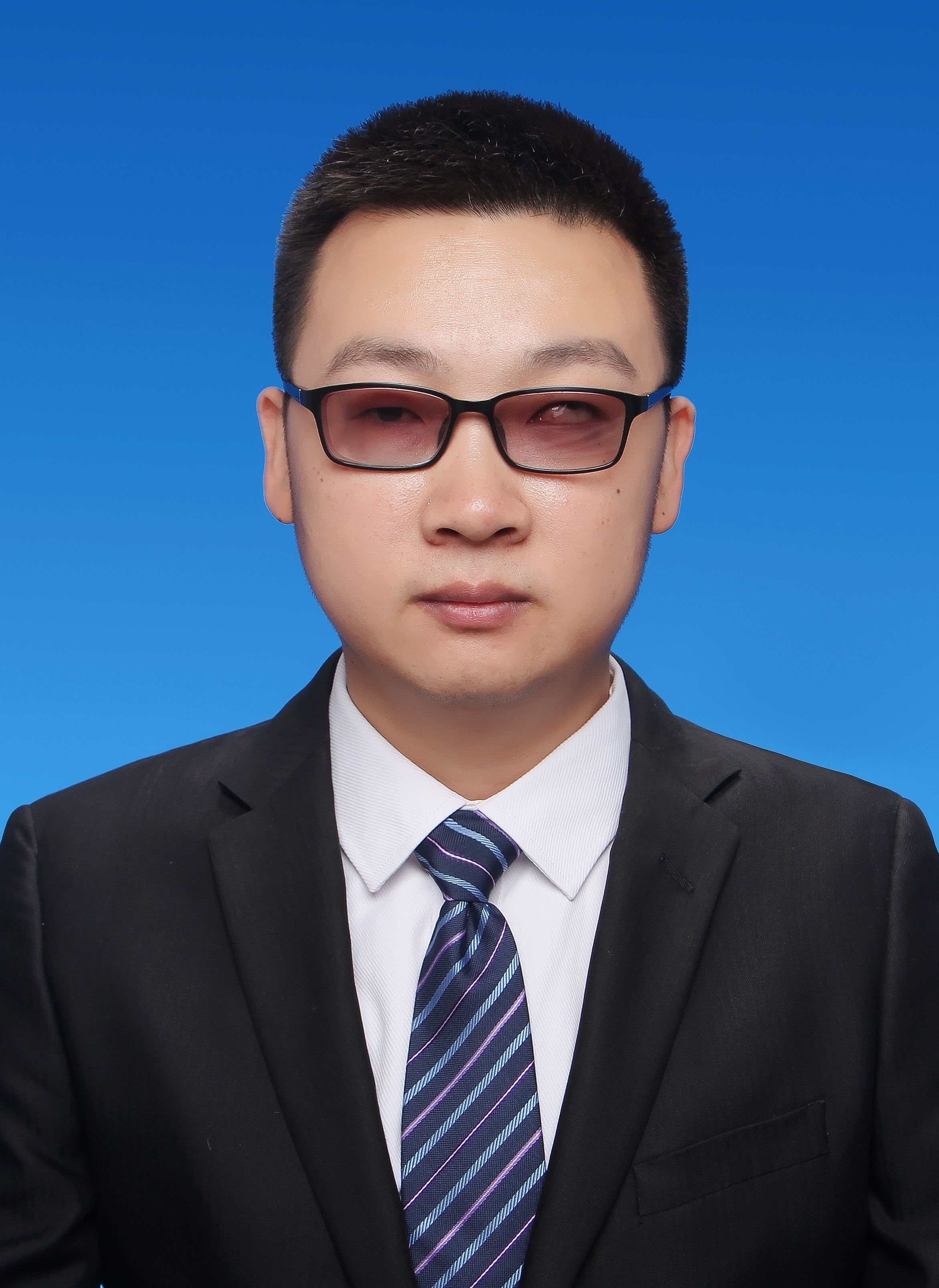}}]{Zhi Chen}
	is now an assistant professor in Southwestern University of Finance and Economics, China. He received his B.S. degree and Ph.D degree in computer science and technology from Sichuan University, China, in 2010 and 2017, respectively. His main research interests are in the area of multiple classifier system, class imbalance learning, feature selection and evolutionary computation. He has published more than 10 peer-reviewed articles on journal like TNNLS, Information Sciences, Knowledge-Based Systems.
\end{IEEEbiography}

\begin{IEEEbiography}[{\includegraphics[width=1in,height=1.25in,clip,keepaspectratio]{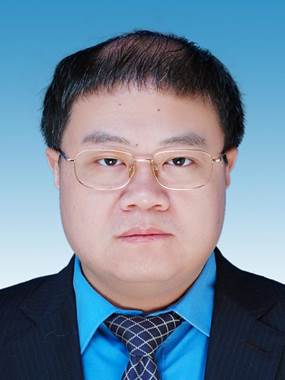}}]{Jiang Duan}
	is now a professor in Southwestern University of Finance and Economics, China. He received the B.S. degree in mechanical engineering from Southwest Jiaotong University, Chengdu, China, in 2001, the M.S. degree from the University of Derby, Derby, U.K., in 2002, and the Ph.D. degree from the University of Nottingham, Nottingham, U.K., in 2006.
	
	Professor Duan is also an expert of the national "Thousand Talents Plan", the winner of the first Sichuan outstanding talent award (the highest talent award of Sichuan province), the winner of the Sichuan youth science and technology award and the standing committee member of Sichuan association for science and technology. Professor Duan has published more than 30 academic papers, won more than 10 international and national patents, and his researches have been funded by about 20 national and provincial funds.
	 
\end{IEEEbiography}

\begin{IEEEbiography}[{\includegraphics[width=1in,height=1.25in,clip,keepaspectratio]{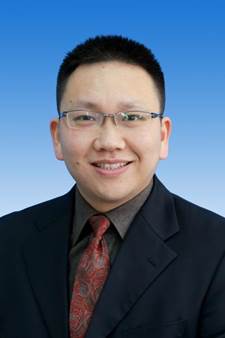}}]{Li Kang}
	is now a professor in Southwestern University of Finance and Economics, China. He received his BS degree in computer science and Ph.D degree in information security from Southwest Jiaotong University, China, in 2004 and 2009, respectively. His main research interests are in the area of public key cryptography,blockchain technology. He has published more than 10 peer-reviewed journal articles and conference papers.
\end{IEEEbiography}

\begin{IEEEbiography}[{\includegraphics[width=1in,height=1.25in,clip,keepaspectratio]{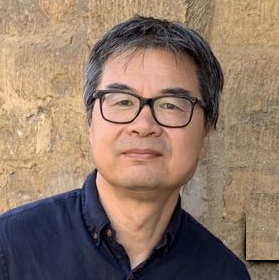}}]{Guoping Qiu}
	is a Distinguished Professor of Information Engineering, Director of Shenzhen University Intelligent Robotics Centre at Shenzhen University, China, and a Chair Professor of Visual Information Processing at the University of Nottingham, Nottingham, UK. He has taught in universities in the UK and Hong Kong and also consulted for multinational companies in Europe, Hong Kong and China. His research interests include image processing, pattern recognition, and machine learning. He is particularly known for his pioneering research in high dynamic range imaging and machine learning based image processing technologies. He has published widely and holds several European and US patents. Technologies developed in his lab have laid the cornerstone for successful spinout companies that are developing advanced digital photography software enjoyed by tens of millions of global users.
\end{IEEEbiography}


%\vfill


\end{document}